\documentclass[10pt,twocolumn,letterpaper]{article}

\usepackage[pagenumbers]{cvpr} 

\usepackage{graphicx}
\usepackage{amsmath}
\usepackage{amssymb}
\usepackage{booktabs}

\usepackage{times}
\usepackage{epsfig}
\usepackage{caption}
\usepackage{multirow,colortbl,tabularx}
\usepackage{xcolor}
\usepackage{lipsum,mdframed}
\usepackage[export]{adjustbox}
\usepackage{wrapfig}

\usepackage[pagebackref,breaklinks,colorlinks]{hyperref}

\usepackage[capitalize]{cleveref}
\crefname{section}{Sec.}{Secs.}
\Crefname{section}{Section}{Sections}
\Crefname{table}{Table}{Tables}
\crefname{table}{Tab.}{Tabs.}

\def\cvprPaperID{103} 
\def\confName{CVPR}
\def\confYear{2022}

\begin{document}

\title{Self-Supervised Material and Texture Representation Learning for Remote Sensing Tasks}

\author{Peri Akiva\\
Rutgers University\\
{\tt\small peri.akiva@rutgers.edu}
\and
Matthew Purri\\
Rutgers University\\
{\tt\small matthew.purri@rutgers.edu}
\and
Matthew Leotta\\
Kitware Inc\\
{\tt\small matt.leotta@kitware.com}
}

\maketitle

\begin{abstract}
\vspace{-1.em}
    Self-supervised learning aims to learn image feature representations without the usage of manually annotated labels. It is often used as a precursor step to obtain useful initial network weights which contribute to faster convergence and superior performance of downstream tasks. While self-supervision allows one to reduce the domain gap between supervised and unsupervised learning without the usage of labels, the self-supervised objective still requires a strong inductive bias to downstream tasks for effective transfer learning. 
    In this work, we present our material and texture based self-supervision method named MATTER (\textbf{MAT}erial and \textbf{TE}xture \textbf{R}epresentation Learning), which is inspired by classical material and texture methods. 
    Material and texture can effectively describe any surface, including its tactile properties, color, and specularity. By extension, effective representation of material and texture can describe other semantic classes strongly associated with said material and texture. MATTER leverages multi-temporal, spatially aligned remote sensing imagery over unchanged regions to learn invariance to illumination and viewing angle as a mechanism to achieve consistency of material and texture representation. We show that our self-supervision pre-training method allows for up to 24.22\% and 6.33\% performance increase in unsupervised and fine-tuned setups, and up to 76\% faster convergence on change detection, land cover classification, and semantic segmentation tasks. 
    %
\end{abstract}

        

\vspace{-1.5em}
\section{Introduction}
Automated understanding of remote sensing imagery has been a long standing goal of the computer vision community. Its broad applicability has driven research and development in construction phase detection \cite{cohen2016rapid}, infrastructure mapping \cite{duan2016towards, lafarge2006automatic, voigt2016global, nayak2002use}, land use monitoring \cite{foody2003remote}, post natural disaster damage assessment \cite{skakun2014flood, van2000remote, gillespie2007assessment}, urban 3D reconstruction \cite{facciolo2017automatic, leotta2019urban}, population migration prediction \cite{chen2020nighttime}, and climate change tracking \cite{rolnick2019tackling}. Most of those methods require some degree of annotation effort, which is often expensive and/or time consuming. Satellite imagery is increasingly plentiful and accessible, with hundreds of satellites collecting images on a daily basis \cite{wmo_oscar_list_of_all_satellites, drusch2012sentinel, torres2012gmes, roy2014landsat}. However, annotating land cover, change, or similar labels often requires domain knowledge and/or extreme attention to detail, as labels in remote sensing imagery cover more numerous and smaller objects seen from unfamiliar view points. As a result,
annotators require more domain expertise compared to standard benchmark datasets such as Pascal VOC \cite{Everingham10}, COCO \cite{lin2014microsoft}, or similar. 


Recent work in self-supervised learning aims to alleviate the requirement of labeled data by either detecting self-applied transformations, such as color or rotation change, or implicit metadata information, such as temporal order or geographical location. Those objectives are often achieved using contrastive learning methods \cite{he2020momentum, khosla2020supervised, chen2020simple}, in which the distance between feature representations of original and transformed images is minimized. More advanced contrastive methods use triplet loss \cite{schultz2004learning, chechik2010large} or quadruplet loss \cite{chen2017beyond} which also include negative examples with which the distance between feature representations is maximized. Despite filling a significant need in the remote sensing domain, these approaches have been yet to be thoroughly investigated. Even methods that utilize contrastive approaches, such as SeCo \cite{Manas_2021_ICCV} and the work of Ayush \etal \cite{ayush2021geography}, which learn
seasonal change invariance or geographic-location consistency, still show weaker transfer-ability to downstream task learning, as demonstrated by inferior performance and convergence speeds shown in Tab. \ref{tab:quant_results_onera}, \ref{tab:quant_results_spacenet}.

Instead, we hypothesize that material and texture have a strong inductive bias to most downstream remote sensing tasks, with pre-training of surface representation to improve performance and convergence speeds (measured in epochs) for those tasks. 
%
%
Consider the task of change detection in remote sensing imagery: when semantic class changes (\ie, soil to building, or forest to soil), change can also be expected in materials and texture, demonstrating the high correlation between material and texture and the change detection task.
%
%
%
%
We show the effectiveness of our self-supervised pre-trained features in both raw and fine-tuned forms, obtaining state-of-the-art (SOTA) performance in change detection (unsupervised and fine-tuned), land cover segmentation (fine-tuned), and land cover classification (fine-tuned). 

Here, we propose a novel self-supervised material and texture representation learning method which is inspired by classical and modern texton filter banks \cite{leung2001representing, zhu2005textons, shotton2006textonboost}. Textons \cite{leung2001representing, julesz1981textons, malik1999textons} refer to the description of micro-structures in images often used to describe material and texture consistency \cite{leung2001representing, wang2004hybrid, cula2001compact, dana2018computational, zhang2015reflectance}. Note that literature has only loosely defined what material, structures, texture, and surface refer to. Here, we define \textit{material} as any single or combination of elements (soil, concrete, vegetation, etc.) corresponding to some multi-spectral signature, \textit{structures} as gradients in intensity, \textit{texture} as spatial distribution of structures, and \textit{surface} as the combination of material and texture. Note that here we define the physical surface, rather than the geometric or algebraic surface, as described by its material and textural properties.
%
By extension, we aim to jointly describe combinations of materials and textures in a single objective. 
For example, within a given image patch, a mixture of grass and concrete should be represented differently than patches with grass or concrete separately. In this example, the grass-concrete mixture may be associated to both grass and concrete material classes. To that end, we learn surface representations that describe the affinity, represented as residuals \cite{jegou2010aggregating}, to all pre-defined surface classes, represented as clusters. 
We achieve this by contrastively learning the similarity between the residuals of multi-temporal, spatially aligned imagery of unchanged regions to obtain consistent material and texture representations, regardless of illumination or viewing angle. This framework acts as a pre-training stage for downstream remote sensing tasks.
%
%

Overall, our contributions are: \textbf{1)} We present a novel material and texture based approach for self-supervised pre-training to generate features with high inductive bias for downstream remote sensing tasks. We propose a texture refinement network to amplify low level features and adapt residual cluster learning to characterize mixed materials and texture patches in a self-supervised, contrastive learning framework. \textbf{2)} We achieve SOTA performance on unsupervised and supervised change detection, semantic segmentation, and land cover classification using our pre-trained network. \textbf{3)} We provide our curated multi-temporal, spatially aligned, and atmospherically corrected remote sensing imagery dataset, collected over unchanged regions used for self-supervised learning. \footnote{Code and dataset will be released upon publication.}

\section{Related Work}
\vspace{-0.2em}
\subsection{Downstream Remote Sensing Tasks}
\vspace{-0.3em}
The main downstream tasks we investigate in this work are change detection, land cover segmentation, and land cover classification. The problem of change detection in satellite imagery has been thoroughly investigated over time \cite{coppin1996digital, peng2019end, chen2020spatial, jiang2020pga, papadomanolaki2019detecting, chen2020dasnet, chen2019deep, sakurada2020weakly}. Notable examples include Daudt \etal \cite{daudt2018fully}, which predicts change by minimizing feature differences at every layer of the network from a given image pair input, and Chen \etal \cite{chen2019deep}, which utilizes a spatial-temporal attention mechanism to detect anomalies in sequences of images. Land cover segmentation and classification have also seen a surge in interest, with growing repositories of annotated datasets \cite{demir2018deepglobe, sumbul2019bigearthnet, alemohammad2020landcovernet, helber2019eurosat, van2018spacenet} and methods \cite{akiva2021h2o, van2018spacenet, rudner2019multi3net, azimi2019skyscapes, tan2020vecroad, gupta2021rescuenet}. H20-Net \cite{akiva2021h2o} synthesizes multi-spectral bands and uses self-sampled points to generate pseudo-ground truth for flood and permanent water segmentation. VecRoad \cite{tan2020vecroad} sets the problem of road segmentation as iterative graph exploration. Multi3Net \cite{rudner2019multi3net} learns fusion of multi-temporal, multi-spectral features from high resolution imagery to jointly predict pixels of floods and building. 



\subsection{Self-Supervision}
\vspace{-0.3em}
In order to effectively utilize large amounts of unlabeled data, recent methods have focused on obtaining good feature representations without explicit annotation efforts. This is done by deriving information from the data itself or learning sub-tasks within data instances without changing the overall objective. The first is often used when high confidence labels can be obtained and trained on, similar to \cite{ayush2021geography, akiva2021h2o}, where the method infers weak supervision about input images through provided meta-data or classical methods. The second, and more common approach, leverages metric learning objectives to learn generalizable features for the same data instance or class. Recent methods involve learning invariance to color and geometric transformations \cite{jing2018self, misra2020self, caron2018deep}, temporal ordering \cite{behrmann2021unsupervised, fernando2017self}, sub-patch relative location prediction \cite{doersch2015unsupervised}, frame interpolation \cite{niklaus2018context}, colorization \cite{zhang2016colorful, deshpande2017learning, larsson2016learning}, patch and background filling \cite{wang2021removing}, and point cloud reconstruction \cite{xie2020pointcontrast}. 

\begin{figure*}[t!]
    \centering
    \includegraphics[width=0.98\textwidth]{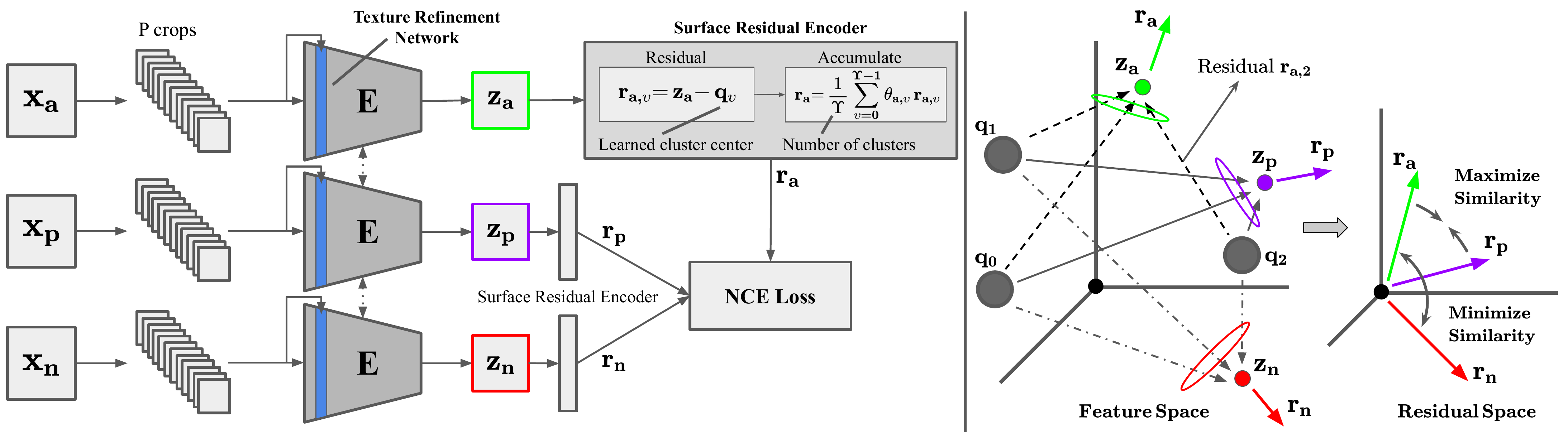}
    \vspace{-0.9em}
    \caption{\textbf{(Left)} \textbf{MATTER}: anchor, positive, and negative images $x_a$, $x_p$, and $x_n$ are densely windowed to $P$ crops which are fed to encoder $E$, and correspond to output features $z_a$, $z_p$, and $z_n$. Crops are also fed to the Texture Refinement Network (shown in blue) which amplifies activation of low-level features to increase their impact in deeper layers. The encoder's output is then fed to the Surface Residual Encoder to generate patch-wise cumulative residuals, which represents affinity between input data and all learned clusters. A residual vector between feature output $z_a$ and cluster $\upsilon$ is denoted as $r_{a,\upsilon}$. Output learned residuals, cluster weights, and number of clusters are noted as $r$, $\theta$, and $\Upsilon$ respectively. \textbf{(Right)} Simplified example of the contrastive objective with $\Upsilon=3$. Residuals from learned clusters are extracted and averaged for all crops as representations of correlation between inputs and all clusters. Best viewed in zoom and color.}
    \label{fig:architecture}
    \vspace{-1.4em}
\end{figure*}

More relevant to the remote sensing domain, SeCo \cite{Manas_2021_ICCV} has taken a step toward utilizing the potential in the abundance of satellite imagery by contrastively learning seasonal invariance as a pre-text self-supervision task. It then fine-tunes the pre-trained network on downstream tasks such as change detection and land cover classification. Ayush \etal \cite{ayush2021geography} also proposes a self-supervised approach enforcing geographical-location consistency as a pre-training objective used for downstream tasks such as land-cover segmentation and classification. While both methods show improved results on benchmark datasets when compared to random weights initialization, we show that their inductive bias is still significantly weaker than that of our material and texture consistency based pre-trained weights, which learn an illumination and viewing angle invariance to achieve consistency of material and texture representation. 
%

\subsection{Material and Texture Identification}
\vspace{-0.3em}
Early material and texture recognition methods relied on hand crafted filter banks, whose combinatorial output are also referred to as textons \cite{leung2001representing}, to encode statistical representations of image patches \cite{leung2001representing, cula2001recognition, dana1999reflectance, zhu1998filters, valkealahti1998reduced, brilakis2005material, brilakis2008shape}. Later works investigated the use of clustering and inter-patch statistics as replacement for pre-defined filter banks \cite{varma2003texture, varma2008statistical}, at the cost of defining the feature space it operates in. Most notable feature spaces include color intensity \cite{chen2009wld}, texture homogeneity \cite{leung2001representing, guo2012discriminative, mao2014active}, multi-resolution features \cite{ojala2002multiresolution, sifre2013rotation}, and feature curvature \cite{liu2010exploring, sharma2012local}. More recent, deep learning approaches have translated the problem of texture representation to focus on explicit identification of materials through texture encoding \cite{zhang2017deep, chen2021deep, zhu2021learning}, differential angular imaging \cite{xue2017differential}, 3D surface variation estimation \cite{degol2016geometry}, auxiliary tactile property \cite{schwartz2013visual}, and radiometric properties estimation such as the bidirectional reflectance distribution function (BRDF) \cite{wang2009material, liu2013discriminative, chen2009robust} and the bidirectional texture function (BTF) \cite{weinmann2014material}. Those methods seek to learn low-level features that are key to material classification and segmentation. Some methods choose to add skip connections \cite{lin2017refinenet, yu2018bisenet, zhu2019asymmetric} to supply low-level features in deep layers, and others choose explicit concatenation of texture related information \cite{schwartz2013visual, li2020improving}. Many of these elements are meant to reduce the receptive field or increase impact of low-level features of the network while keeping it sufficiently deep. FV-CNN \cite{cimpoi2015deep} aims to generate texture descriptions of densely sampled windows. Since the features describe regions removed from global spatial information, it explicitly constrains the receptive field of the network to the size of the window. DeepTEN \cite{zhang2017deep} learns residual representations of material images in an end-to-end pipeline using material labels. Our approach combines elements from FV-CNN and DeepTEN in two ways. First, we densely sample windows and refine low level features as receptive field constraints. Then, we contrastively learn implicit surface residual representations without the usage of material labels or auxiliary information. To our knowledge, we are the first to employ self-supervised material and texture based objectives for pre-training steps.


%
\section{Methodology}
\begin{figure}[t!]
    \centering
    \includegraphics[width=0.99\columnwidth]{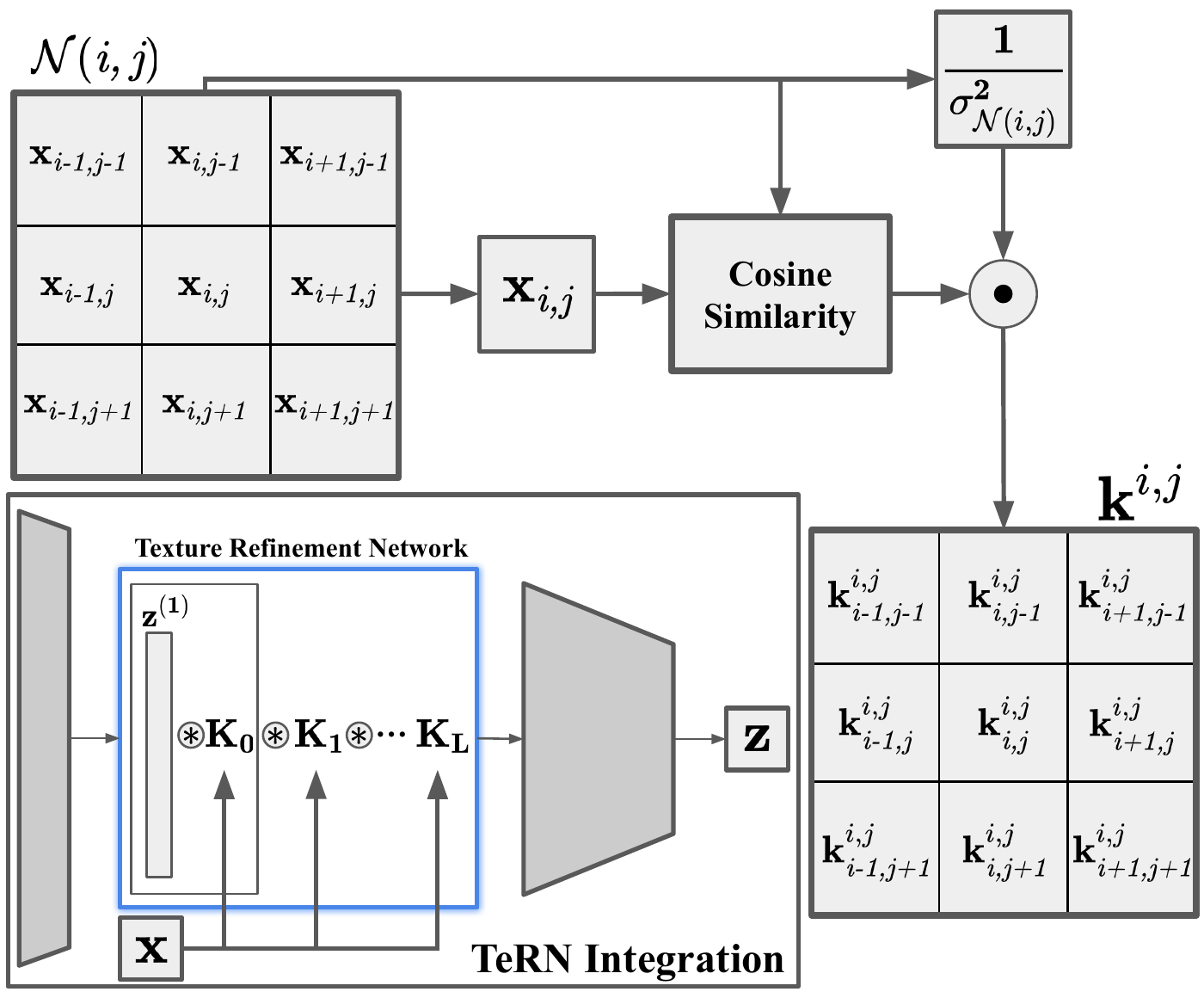}
    \vspace{-0.5em}
    \caption{\textbf{Texture Refinement Network (TeRN)} assigns convolution weights based on the cosine similarity of the kernel's center pixel and its neighbors, divided by the standard deviation of that kernel. We then convolve features $z^{(1)}$ to refine low-level features essential for texture and material centric learning. The symbols $\circledast$ and $\odot$ correspond to convolution and element-wise multiplication operations. Best viewed in zoom.}
    \label{fig:tern_network_arch}
    \vspace{-1.5em}
\end{figure}

\vspace{-0.3em}
The goal of \textbf{MAT}erial and \textbf{TE}xture \textbf{R}epresentation Learning (MATTER) is to learn a feature extractor that generates illumination and viewing angle invariant material and texture representations from given multi-temporal satellite imagery sampled over unchanged regions. To this end, to train our model, we utilize our self-collected dataset described in Sec. \ref{sec:pre_training_subsec}, which samples multi-temporal imagery of rural and remote regions, in which little to no change is assumed between every consecutive pair of sampled images. 
See Fig.~\ref{fig:tern_network_arch} for an overview of our approach.

Given an anchor reference image $x_a \in \mathcal{R}^{B\times H \times W}$ sampled over an unchanged region, we obtain a positive, temporally succeeding image $x_p \in \mathcal{R}^{B\times H \times W}$ over the same region, and a negative image $x_n \in \mathcal{R}^{B\times H \times W}$ sampled over a different region. $B$, $H$, and $W$ correspond to number of channel bands, height, and width of input images. We tile all images into $P$ equally sized, corresponding patches of size $h \times w$, with spatially aligned reference and positive patches, $c_a$ and $c_p$, and negative patches, $c_n$, randomly sampled from regions other than the reference region. The usage of densely sampled crops aims to restrict the receptive field by removing features from the global spatial context, and to prevent the model from learning higher level features ineffective in describing surfaces. We study the effects of receptive field variation in Sec. \ref{sec:ablation}.

To learn material and texture centric features, we present the Texture Refinement Network (TeRN) (Sec. \ref{sec:tern}), and patch-wise Surface Residual Encoder (Sec. \ref{sec:tex_residual}). TeRN aims to amplify the activation of lower level features essential for texture representation (as seen in Fig. \ref{fig:tern_qualitative}), and Surface Residual Encoder is our patch-wise adaptation of Deep-TEN \cite{zhang2017deep} to learn surface-based residual representations. We train our network to minimize the feature distance of positive patch pairs, $c_a$ and $c_p$, and maximize the feature distance of negative patch pairs, $c_a$ and $c_n$, where the features are the learned residual representations. For our learning objective, we use the Noise Contrastive Estimation loss \cite{oord2018representation}: 
%
%
\vspace{-1.8em}

\begin{equation}
    \begin{aligned}
        \mathcal{L}_{NCE}= -\mathbb{E}_C \Biggr[log\frac{\text{exp}(f(c_a)\cdot f(c_p))}{\sum\limits_{c_j \in C} \text{exp}(f(c_a)\cdot f(c_j))}\Biggr],
    \end{aligned}
    \vspace{-0.7em}
\end{equation}
%
%
where $f(c_j)$ is the output features of input patch $c_j$, and $C$ is the set of positive and negative patches. 
%

\subsection{Texture Refinement Network}
\label{sec:tern}
\begin{figure}[t!]
\setlength\tabcolsep{1pt}
\def\arraystretch{0.5}
\centering
\begin{tabular}{@{\hspace{0.3em}}ccc@{}}
    \includegraphics[width=0.26\linewidth]{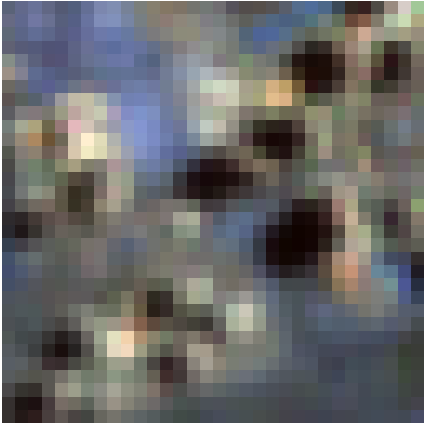}&
    \includegraphics[width=0.26\linewidth]{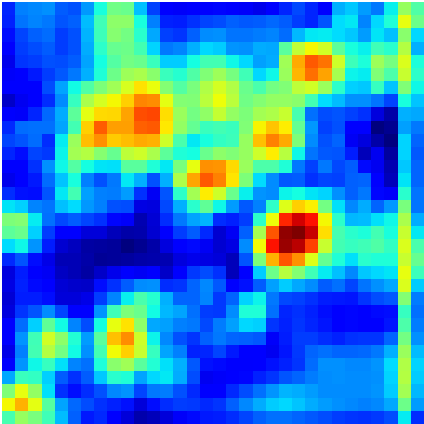} & 
    \includegraphics[width=0.26\linewidth]{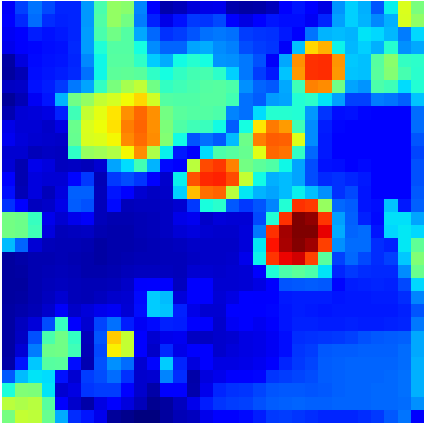} \\
    \includegraphics[width=0.26\linewidth]{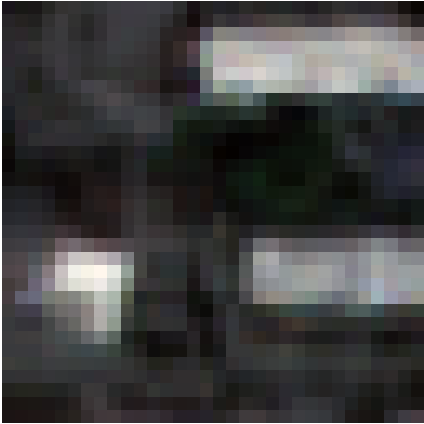} &
    \includegraphics[width=0.26\linewidth]{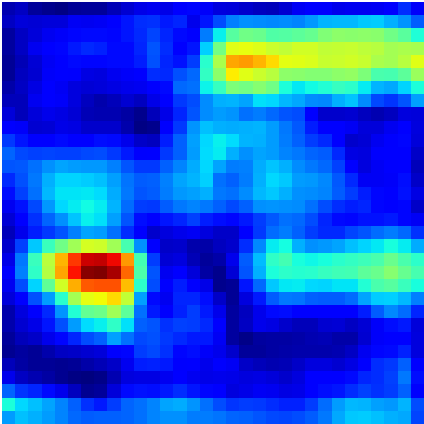} & 
    \includegraphics[width=0.26\linewidth]{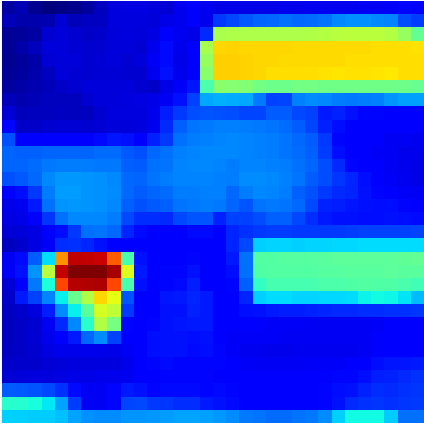} \\
    Input Image & 
    Raw Features & 
    Refined Features
\end{tabular}
\vspace{-0.45em}
\caption{\textbf{Qualitative results} of our Texture Refinement Network (TeRN). It can be seen that similar textured pixels obtain similar feature activation intensity in the refined output. Notice how the building in the second row obtains similar activation throughout the concrete building pixel locations compared to the raw features output. Best viewed in zoom and color.}
\vspace{-1.6em}
\label{fig:tern_qualitative}
\end{figure}

Capturing texture details is difficult in low resolution images, and is especially challenging when considering satellite images that have low contrast. As a result, texture will be less visible and have less impact on the final extracted features. We address this challenge by using our Texture Refinement Network (TeRN) to refine lower level texture features to increase their impact in deeper layers. TeRN utilizes the recently introduced pixel adaptive convolution layer \cite{su2019pixel}, in which the convolution kernel weights are a function of the features locally confined by the kernel. Here, our kernel considers the corresponding local pixels in the original image as follows: given kernel $k^{i,j}$ centered at location $(i,j)$, we calculate the cosine similarity between pixel $x_{i,j}$ and all of its neighboring pixels $\mathcal{N}(i,j)$.
The output matrix is then divided by the squared standard deviation of all pixels within $\mathcal{N}(i,j)$, noted as $\sigma_{\mathcal{N}(i,j)}$. 
\vspace{-0.3em}
\begin{equation}
    k^{i,j} = - \frac{1}{\sigma_{\mathcal{N}(i,j)}^2}\frac{x_{i,j} \cdot x_{p,q}}{||x_{i,j}||_2 \cdot ||x_{p,q}||_2}, \hspace{0.2em} \forall \hspace{0.2em} p,q \in \mathcal{N}(i,j).
    \vspace{-0.3em}
\end{equation}
The output matrix of those operations describes both the similarity of the center pixel to its surroundings, and the intensity gradients within the kernel. As previously defined, texture is the spatial distribution of structures, which are represented as intensity gradients. Since we want to emphasize texture, we explicitly polarize the feature activation in regions with high variance or low similarity, with kernel weights decreasing with high variance and/or low cosine similarity. When convolved over our low level features, it highlights edges, and encourages representation consistency for pixels with similar material signatures, as seen in figure \ref{fig:tern_qualitative}. The described operation constitutes a single kernel location of a single refinement layer. We define a single refinement layer, $K$, when the operation is repeated for all image locations. We construct an $L$-layer refinement network, where each layer is able to utilize different kernel sizes, dilations, and strides. Since the network has deterministicly defined weights, it does not have learned parameters. A base TeRN kernel, and its integration in the overall network, are visually depicted in Fig. \ref{fig:tern_network_arch}, and sample refined features in Fig. \ref{fig:tern_qualitative}.

\subsection{Learning Consistency of Surface Residuals}
\label{sec:tex_residual}
The task of residual encoding is tightly related to classical k-means clustering \cite{lloyd1982least} and bag-of-words \cite{joachims1998text}, in which some hard cluster assignment is learned based on the data instance proximity to cluster centers. Given the cluster centers, the residual is calculated as the distance of any data instance from its corresponding cluster center. In practice, we can use the residual to measure how similar a given data instance is to its assigned cluster, and all other clusters. 
Our method adapts the work presented in Deep-TEN \cite{zhang2017deep} to learn patch-wise residual encodings without explicit hand-crafted clustering through a differentiable pipeline. Traditionally, in Deep-TEN \cite{zhang2017deep} and other classical and deep clustering methods \cite{cheng1995mean, macqueen1967some, he2020momentum, caron2018deep, chen2020simple}, the objective is to cluster image-wise inputs to corresponding class-wise cluster centers. In contrast, as we employ a patch-wise approach. A given patch containing some material and texture may be associated with multiple clusters (\ie if a patch captures multiple material elements), so it requires a soft representation depicting an affinity to all learned clusters, and not only to its closest cluster. 
%

Consequently, we learn residuals of small patches and enforce multi-temporal consistency between corresponding patch residuals, imposing similarity of cluster affinity. Given an output feature vector $z^{1\times D}_i$ for some crop $c_i$, and a set of $\Upsilon$ learned cluster centers $Q = \{q_0, q_2, ..., q_{\Upsilon-1}\}$, each of shape $1 \times D$, we can find the residual corresponding to the feature vector $z_i$ and learned cluster center $q_{\upsilon}$ using $r^{1 \times D}_{i,\upsilon} = z_i - q_\upsilon$. We repeat this for all cluster centers and take the weighted average of residuals from each cluster to obtain the cumulative residual vector, 
\vspace{-0.8em}
\begin{equation}
r_i = \frac{1}{\Upsilon}\sum_{\upsilon=0}^{\Upsilon-1} \theta_{i,\upsilon} r_{i,\upsilon},
\vspace{-0.6em}
\end{equation}
with learned cluster weight $\theta_{\upsilon}$. By combining the residuals of a given crop, we represent its affinity with all learned clusters. When maximizing or minimizing similarity between residuals, we effectively enforce a consistent cluster affinity between input crops. 



\vspace{-0.3em}
\section{Experiments}
\vspace{-0.2em}
\subsection{Self-Supervised Pre-Training}
\label{sec:pre_training_subsec}
\vspace{-0.3em}
\paragraph{Pre-Training Dataset.} To train our self-supervised task, we collect a large amount of freely available, orthorectified, atmospherically corrected Sentinel-2 imagery of regions with limited human development. Regions of interest were manually selected to cover a variety of climates. Given spatial and temporal ranges, we use the PyStac library \cite{pystac} to fetch imagery from the AWS Sentinel-2 catalog closest to our points of interest. Imagery within the spatial-temporal constraints containing over 20\% cloud cover and less than 80\% data coverage were removed. A maximum of 100 images meeting these constraints were collected per region. The collected images were divided into 14,857, 1096$\times$1096 $px^2$ sized tiles for training. The resultant dataset contains 27 regions of interest spanning 1217 $km^2$ over three years. We provide all points of interest (Lat., Long.) in the supplementary material, and will release the dataset upon publication.
\begin{table}[t!]
\centering
\resizebox{0.47\textwidth}{!}{%
\begin{tabular}{l@{\hspace{0.8em}}c@{\hspace{0.8em}}c@{\hspace{0.8em}}c@{\hspace{0.8em}}c}
\toprule
Dataset & & \multicolumn{3}{c}{OSCD \cite{8518015}} \\ \midrule
Method & Sup. & Precision (\%) & Recall (\%) & F-1 (\%)\\ \midrule
\multicolumn{5}{l}{\centering \textit{Full Supervision}} \\ \midrule
U-Net \cite{ronneberger2015u} (random)  & $\mathcal{F}$ & \textbf{70.53} & 19.17  & 29.44   \\
U-Net \cite{ronneberger2015u} (ImageNet)  & $\mathcal{F}$ & 70.42  & 25.12  & 36.20  \\
MoCo-v2 \cite{he2020momentum}  & $\mathcal{S}$ + $\mathcal{F}$ & 64.49 & 30.94 & 40.71  \\
SeCo \cite{Manas_2021_ICCV}  & $\mathcal{S}$ + $\mathcal{F}$ & 65.47 & 38.06 & 46.94 \\
DeepLab-v3 \cite{chen2017deeplab} (ImageNet) & $\mathcal{F}$ & 51.63 & 51.06  & 53.04   \\
Ours (fine-tuned) & $\mathcal{S}$ + $\mathcal{F}$&  61.80 & \textbf{57.13} & \textbf{59.37} \\
\midrule    
\multicolumn{5}{l}{\centering \textit{Self-Supervision only}} \\ \midrule
VCA \cite{malila1980change} & $\mathcal{S}$ & 9.92 & 20.77 & 13.43 \\
MoCo-v2 \cite{he2020momentum}  & $\mathcal{S}$ & 29.21 & 11.92 & 16.93 \\
SeCo \cite{Manas_2021_ICCV}  & $\mathcal{S}$ & \textbf{74.70} & 15.20 & 25.26 \\
Ours  & $\mathcal{S}$ & 37.52 & \textbf{72.65} & \textbf{49.48} \\
\bottomrule
\end{tabular}%
}
\vspace{-0.5em}
\caption{Precision, recall, and F-1 (\%) accuracies (higher is better) of the "change" class on Onera Satellite Change Detection (OSCD) dataset validation set \cite{8518015}. $\mathcal{F}$, and $\mathcal{S}$ represent full and self-supervision respectively. $\mathcal{S}$ + $\mathcal{F}$ refer to self-supervised pre-training followed by fully supervised fine-tuning. Random and ImageNet denote the type of backbone weight initialization that method uses.}
\vspace{-1.8em}
\label{tab:quant_results_onera}
\end{table}
\vspace{-2.2em}
\paragraph{Implementation Details.} We adopt a standard ResNet-34 backbone, with TeRN inserted after the first layer, and the Surface Residual Encoder as the output layer. TeRN is constructed with 10 blocks, each containing three layers of kernel size $3 \times 3$ and dilations of 1-1-2. For the Surface Residual Encoder, we use $\Upsilon = 64$. We use training patch size of $7 \times 7$, batch size of 32, learning rate of $0.01$, momentum of $0.6$, and weight decay of $0.001$ for training. For the Noise Contrastive Estimation loss, we use a temperature scaling of $0.05$. We pre-train the network for 110,000 iterations or until convergence. Note that the self-supervised baselines SeCo \cite{Manas_2021_ICCV} and Ayush \etal \cite{ayush2021geography} use 1 million and 543,435 images respectively for pre-training, while we use only 14,857 images.
\vspace{-0.6em}
\subsection{Change Detection}
\label{sec:change_detection}
\begin{figure*}[t!]
\setlength\tabcolsep{1pt}
\def\arraystretch{0.5}
\centering
\begin{tabular}{ccccccccc}
    \includegraphics[width=0.133\linewidth]{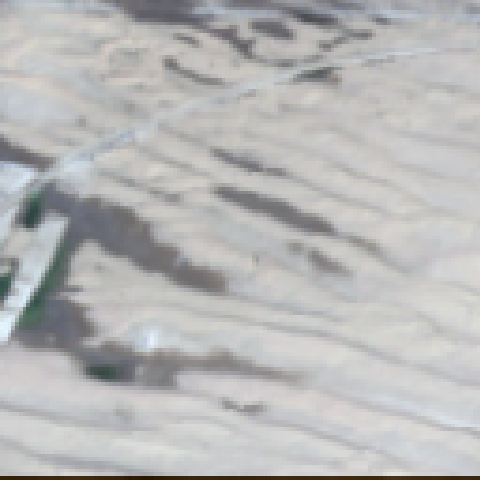} &
    \includegraphics[width=0.133\linewidth]{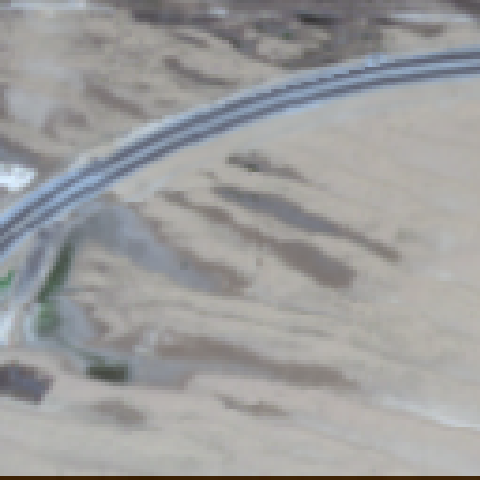} &
    \includegraphics[width=0.133\linewidth]{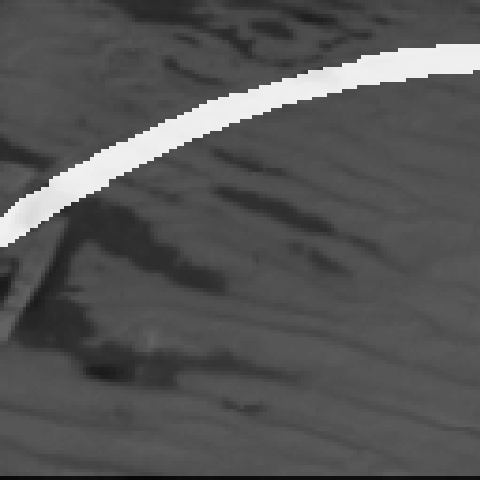} &
    \includegraphics[width=0.133\linewidth]{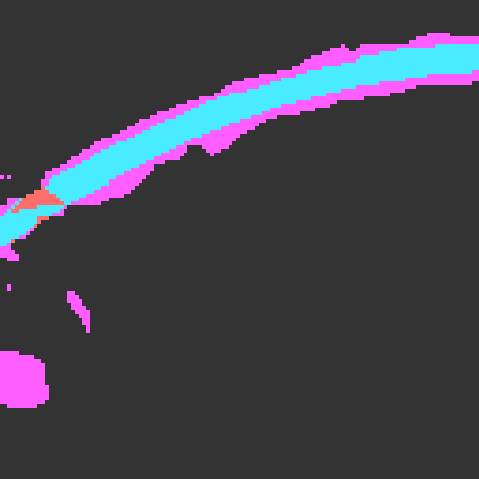} &
    \includegraphics[width=0.133\linewidth]{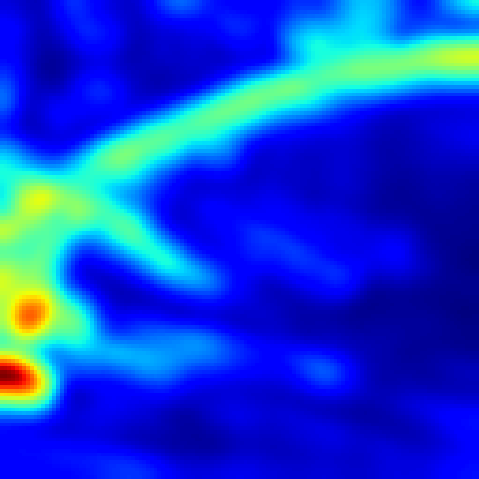} &
    \includegraphics[width=0.133\linewidth]{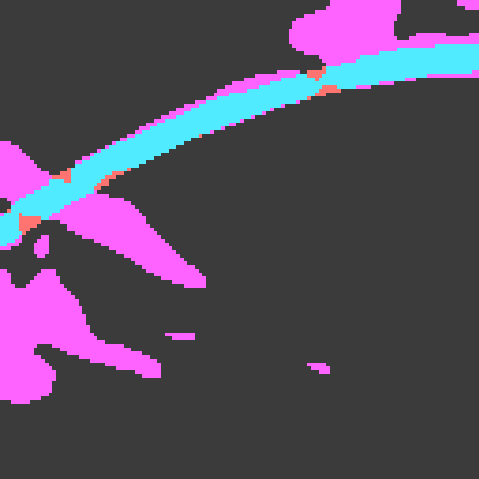} &
    \includegraphics[width=0.133\linewidth]{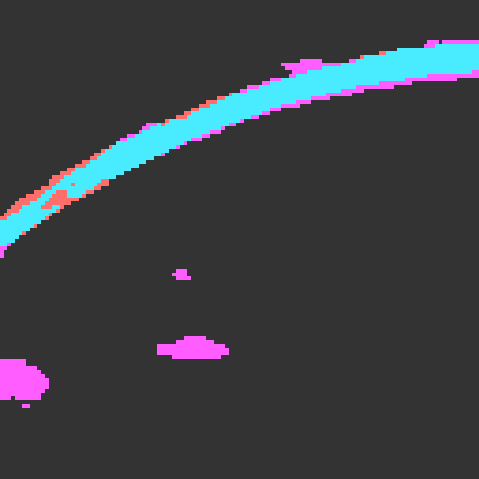} \\
    \includegraphics[width=0.133\linewidth]{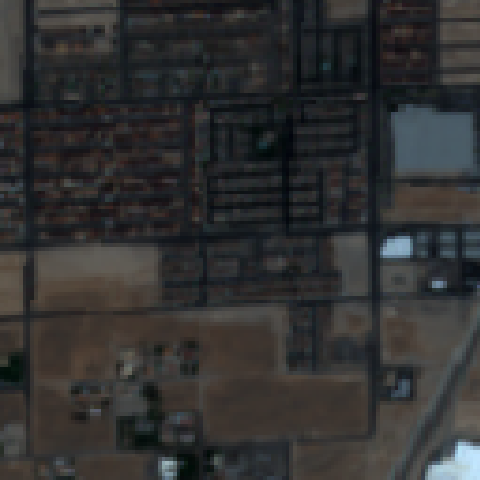} &
    \includegraphics[width=0.133\linewidth]{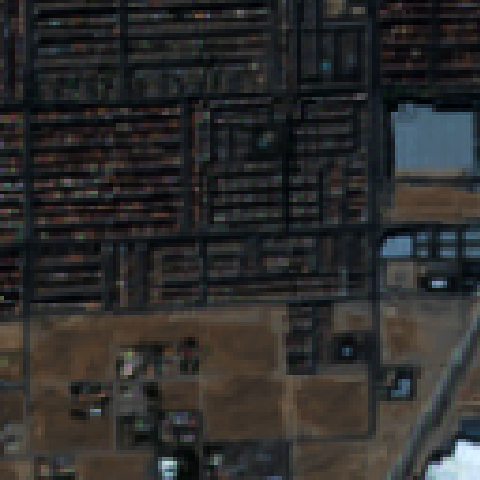} &
    \includegraphics[width=0.133\linewidth]{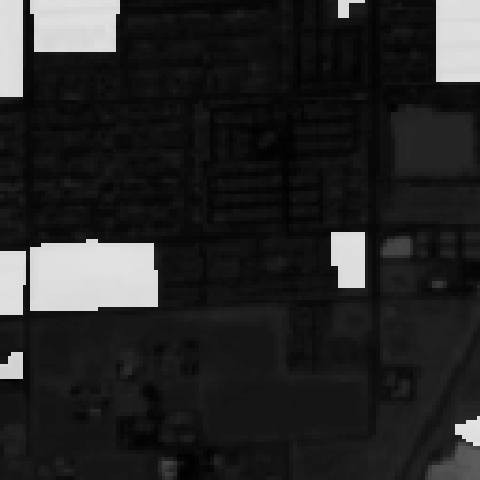} &
    \includegraphics[width=0.133\linewidth]{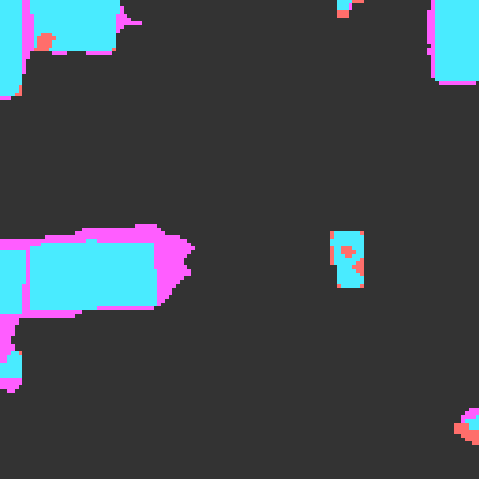} &
    \includegraphics[width=0.133\linewidth]{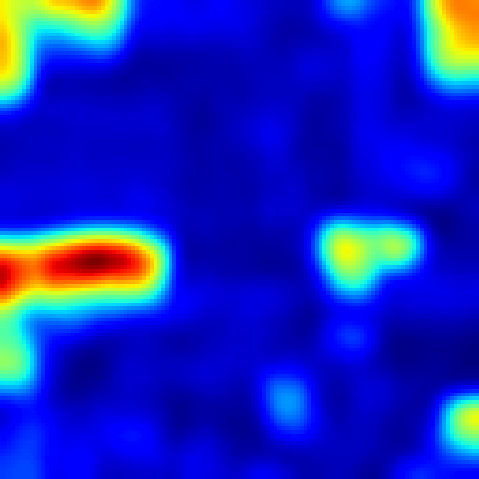} &
    \includegraphics[width=0.133\linewidth]{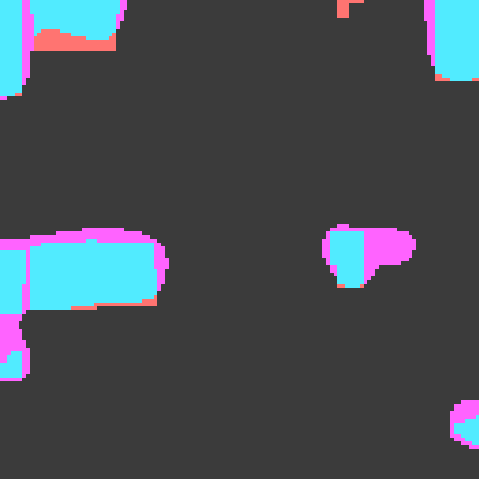} &
    \includegraphics[width=0.133\linewidth]{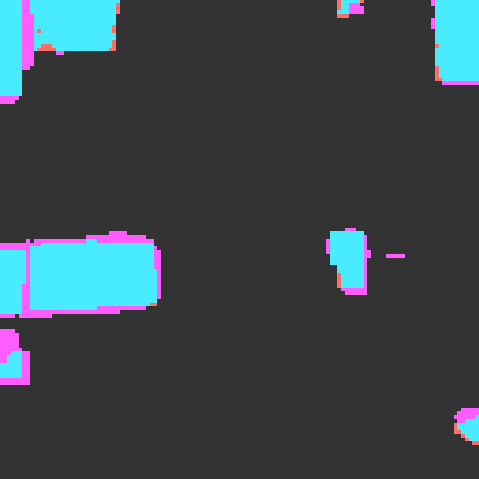} \\
    \includegraphics[width=0.133\linewidth]{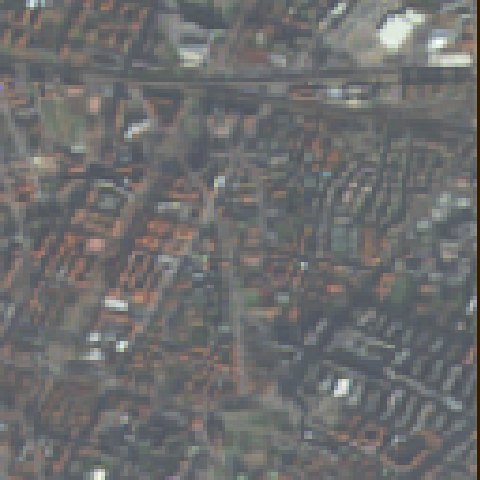} &
    \includegraphics[width=0.133\linewidth]{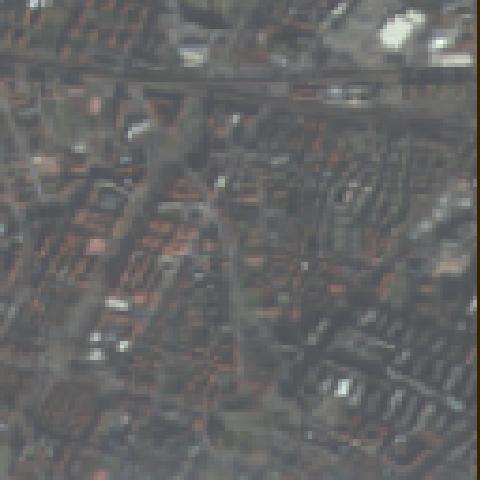} &
    \includegraphics[width=0.133\linewidth]{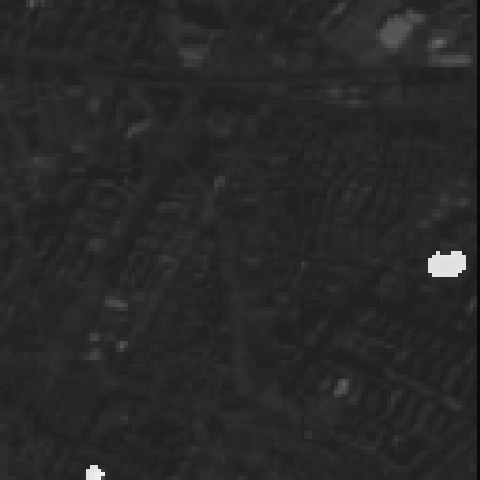} &
    \includegraphics[width=0.133\linewidth]{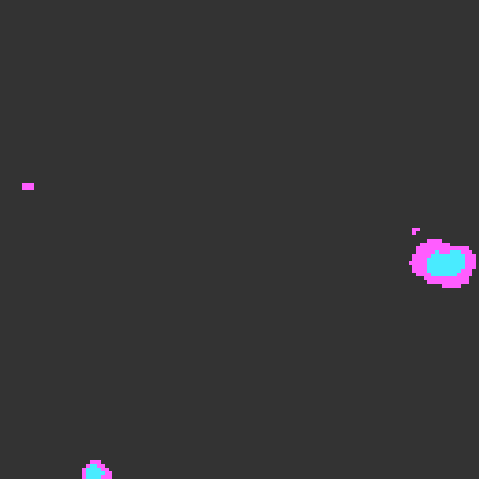} &
    \includegraphics[width=0.133\linewidth]{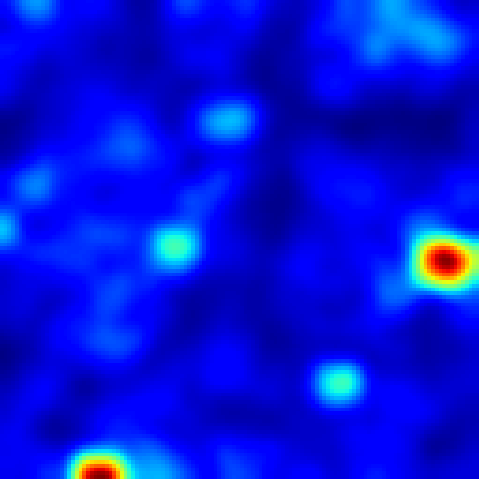} &
    \includegraphics[width=0.133\linewidth]{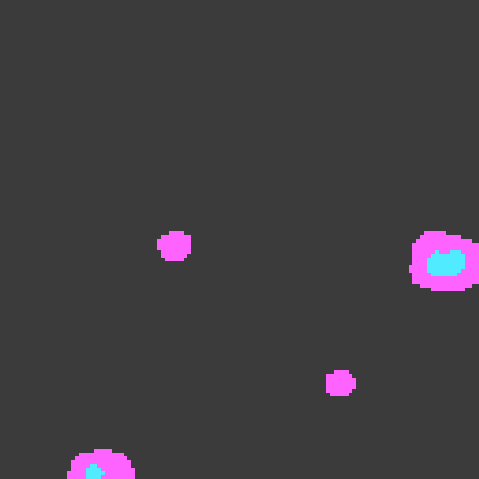} &
    \includegraphics[width=0.133\linewidth]{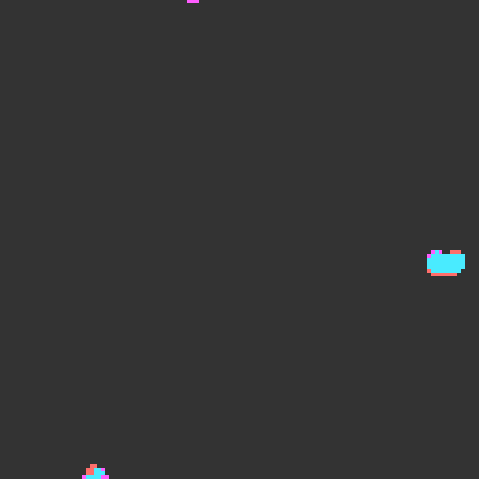} \\
    \includegraphics[width=0.133\linewidth]{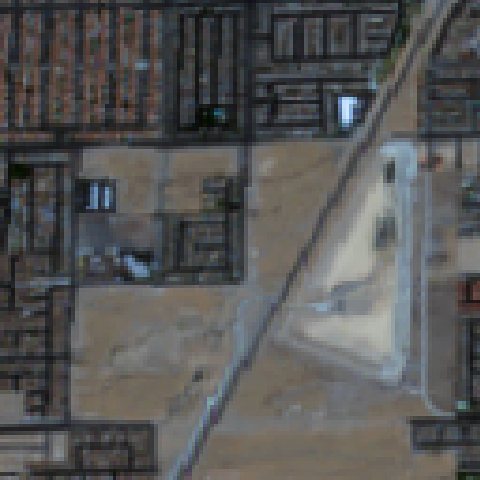} &
    \includegraphics[width=0.133\linewidth]{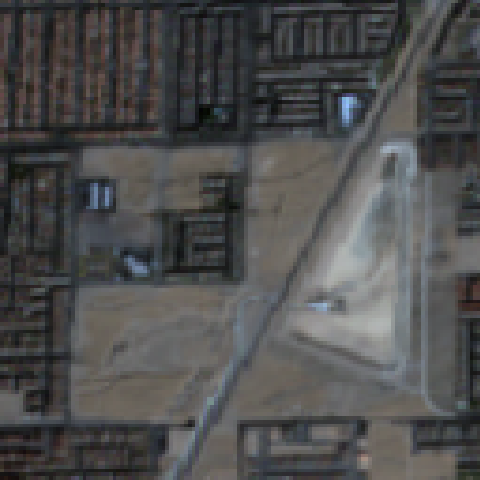} &
    \includegraphics[width=0.133\linewidth]{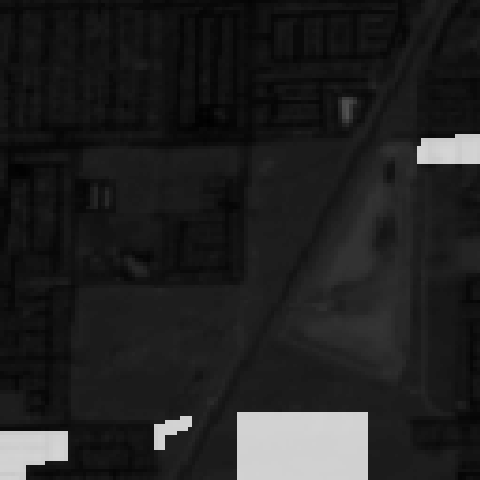} &
    \includegraphics[width=0.133\linewidth]{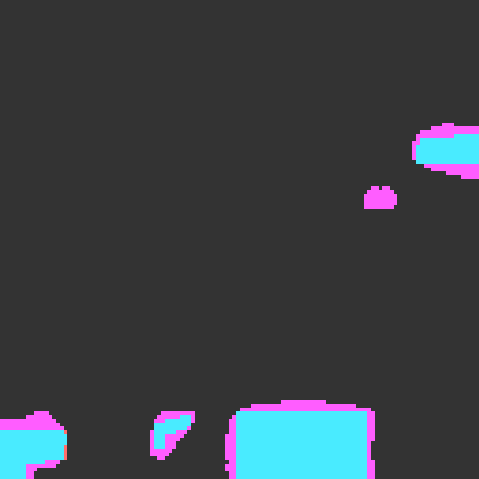} &
    \includegraphics[width=0.133\linewidth]{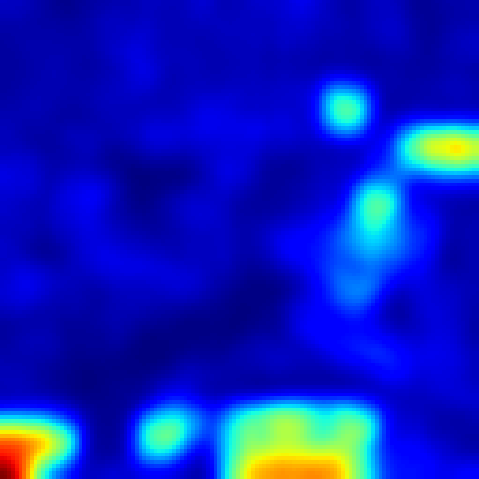} &
    \includegraphics[width=0.133\linewidth]{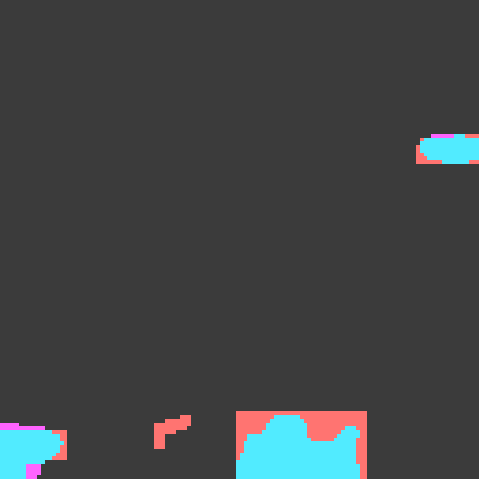} &
    \includegraphics[width=0.133\linewidth]{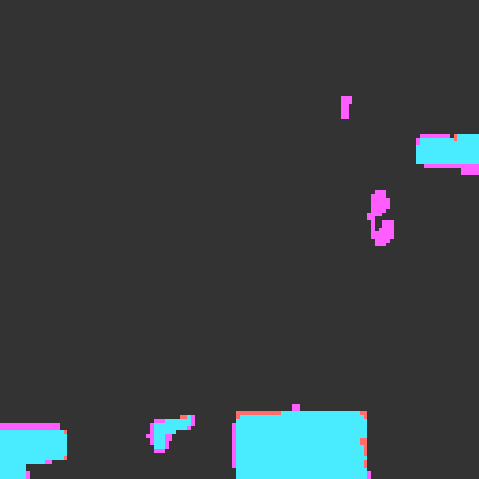} \\
    Image 1 & 
    Image 2 & 
    Ground Truth  & 
    DeepLab \cite{chen2017deeplab} &
    \multicolumn{1}{m{2.3cm}}{\centering Difference of Residuals} &
    Ours ($\mathcal{S}$)&
    Ours ($\mathcal{S}+\mathcal{F}$) 
\end{tabular}
\vspace{-0.7em}
\caption{\textbf{Qualitative results} of our method on Onera Satellite Change Detection (OSCD) dataset \cite{8518015}. It can be seen that our self-supervised alone is capable of detecting change only by inferring on the change of material and texture. The fine-tuned model is able to utilize the pre-trained material and texture based weights and achieve significantly better results than models with ImageNet weight initialization. Cyan, magenta, gray, and red colors represent true positive, false positive, true negative, and false negative respectively. Best viewed in zoom and color.}
\vspace{-1.7em}
\label{fig:qual_results}
\end{figure*}

\vspace{-0.3em}
\paragraph{Implementation Details.} This task is evaluated on the Onera Satellite Change Detection (OSCD) dataset \cite{8518015}, and performed in two ways: self-supervised, and supervised fine-tuning. The self-supervised approach utilizes \textit{only} the pre-trained backbone to extract patch-wise residual features from both images, with each $9 \times 9$ patch representing its center pixel. We calculate the euclidean distance as a change metric between corresponding residual features, which are thresholded using Otsu thresholding \cite{otsu1979threshold} to predict change pixels when residual distance is large. For the fine-tuned approach, we use image-wise inputs to a DeepLab-v3 \cite{chen2017deeplab} with skip-connections network with our pre-trained backbone, fine-tuning the decoder for 30 epochs while freezing the backbone's weights. We use channel-wise concatenations of image pairs as input to the network, with the output features optimized using the cross entropy loss and ground truth change masks. For evaluation, we report precision, recall, and F-1 score of the ``change" class in Tab. \ref{tab:quant_results_onera}. We use batch-size of $32$, learning rate of $0.001$, momentum of $0.6$, and weight decay of $0.001$. For the self-supervised baseline methods, we use the publicly available model weights and follow the same previously described self-supervised change prediction pipeline. The fully-supervised baselines follow the same steps as our fine-tuned approach, without the pre-trained weight initialization.
%
%
%
%
\vspace{-1.2em}
\paragraph{Results Discussion.} In Tab. \ref{tab:quant_results_onera} and Fig. \ref{fig:qual_results} we compare our method to SOTA baselines for both self-supervised and fine-tuned approaches. We present common semantic segmentation networks initialized with weights that are random, or pre-trained with ImageNet \cite{krizhevsky2012imagenet}, MoCo-v2 \cite{he2020momentum}, and SeCo \cite{Manas_2021_ICCV}. We hypothesized that change in material and texture corresponds to actual change in the scene. Hence by learning good material and texture representation and comparing representations of image pairs, we can reliably locate regions of change. As evident by Tab. \ref{tab:quant_results_onera}, our self-supervised approach learns sufficiently good material and texture representation to outperform other fine-tuned methods, surpassing self-supervised SeCo by 24.22\%, and fine-tuned SeCo by 2.08\%. When considering our fine-tuned method, we outperform our baselines even further, with 12.43\% performance increase compared to our self-supervision based baseline, and 6.33\% performance increase compared to the fully supervised baseline. Additionally, we show that the inductive bias of material and texture representation to the task of change detection is significant as evidenced by the quicker convergence speed (measured in epochs), with our method converging within only 30 epochs, compared to 100 epochs reported by SeCo.


\begin{table}[t!]
\centering
\resizebox{0.47\textwidth}{!}{%
\begin{tabular}{l@{\hspace{0.8em}}c@{\hspace{0.8em}}c@{\hspace{0.8em}}c}
\toprule
Dataset & \multicolumn{3}{c}{BigEarthNet \cite{sumbul2019bigearthnet}} \\ \midrule
Method & Sup. & Fine-Tune Epochs & mAP (\%) \\ \midrule
Inception-v2 \cite{szegedy2016rethinking} & $\mathcal{F}$ & - & 48.23  \\
InDomain \cite{neumann2019domain} & $\mathcal{S}$ + $\mathcal{F}$ & 90 & 69.70 \\
S-CNN \cite{sumbul2019bigearthnet} & $\mathcal{F}$ & - & 69.93  \\
ResNet-50 \cite{he2016deep} (random) & $\mathcal{F}$ & - & 78.98  \\
ResNet-50 \cite{he2016deep} (ImageNet) & $\mathcal{F}$ & - & 86.74  \\
MoCo-v2 \cite{he2020momentum}  & $\mathcal{S}$ + $\mathcal{F}$ & 100 & 86.05  \\
SeCo \cite{Manas_2021_ICCV}  & $\mathcal{S}$ + $\mathcal{F}$ & 100 & 87.81 \\
Ours (fine-tuned) & $\mathcal{S}$ + $\mathcal{F}$ &  24 & \textbf{87.98}  \\
\bottomrule
\end{tabular}%
}
\vspace{-0.4em}
\caption{Mean average precision accuracy (higher is better) on BigEarthNet land cover multi-label classification dataset validation set \cite{sumbul2019bigearthnet}. $\mathcal{F}$, and $\mathcal{S}$ represent full and self-supervision respectively. $\mathcal{S}$ + $\mathcal{F}$ refer to self-supervised pre-training followed by fully supervised fine-tuning.}
\vspace{-1.6em}
\label{tab:quant_results_bigearthnet}
\end{table}

\subsection{Land Cover Classification}
\vspace{-0.3em}
\paragraph{Implementation Details.} We evaluate our pre-trained backbone on the BigEarthNet \cite{sumbul2019bigearthnet} dataset for the task of multi-label land cover classification. The dataset provides 590,326 multi-spectral images of size $120 \times 120$ annotated with multiple land-cover labels, split into train and validation sets (95\%/5\%). We fine-tune a classifier head added to our frozen pre-trained backbone network for 24 epochs using given ground truth labels. We use SGD optimizer, batch-size of $128$, learning rate of $0.0005$, momentum of $0.6$, and weight decay of $0.001$. For performance, we report the mean average precision of all classes (19).
\vspace{-1.0em}
\paragraph{Results Discussion.} Tab. \ref{tab:quant_results_bigearthnet} reports the mean average precision performance of baseline and our methods after fine-tuning. While our method only outperforms our baseline by 0.18\%, we note that our method converges within 24 epochs, which is significantly faster than our best-performing baseline which reports convergence within 100 epochs.

\begin{table}[t!]
\centering
\resizebox{0.47\textwidth}{!}{%
\begin{tabular}{l@{\hspace{0.8em}}c@{\hspace{0.8em}}c@{\hspace{0.8em}}c}
\toprule
Dataset & \multicolumn{3}{c}{SpaceNet \cite{van2018spacenet}} \\ \midrule
Method & Sup. & Fine-Tune Epochs & mIoU (\%) \\ \midrule
DeepLab-v3 \cite{chen2017deeplab} (random)  & $\mathcal{F}$ & - & 69.44  \\
DeepLab-v3 \cite{chen2017deeplab} (ImageNet)  & $\mathcal{F}$ & - & 72.22   \\
MoCo-v2 \cite{he2020momentum}  & $\mathcal{S}$ + $\mathcal{F}$ & 100 & 78.05 \\
Ayush \etal \cite{ayush2021geography} & $\mathcal{S}$ + $\mathcal{F}$ & 100 & 78.51  \\
Ours (fine-tuned) & $\mathcal{S}$ + $\mathcal{F}$ & 24 & \textbf{81.12}  \\
\bottomrule
\end{tabular}%
}
\vspace{-0.4em}
\caption{Mean intersection over union (higher is better) on SpaceNet building segmentation dataset validation set \cite{van2018spacenet}. $\mathcal{F}$, and $\mathcal{S}$ represent full and self-supervision respectively. $\mathcal{S}$ + $\mathcal{F}$ refer to self-supervised pre-training followed by fully supervised fine-tuning. Random and ImageNet denote the type of backbone weight initialization that method uses.}
\vspace{-0.9em}
\label{tab:quant_results_spacenet}
\end{table}
\begin{figure}[t!]
\setlength\tabcolsep{1pt}
\def\arraystretch{0.5}
\centering
\begin{tabular}{@{}cccc@{}}
    \includegraphics[width=0.24\linewidth]{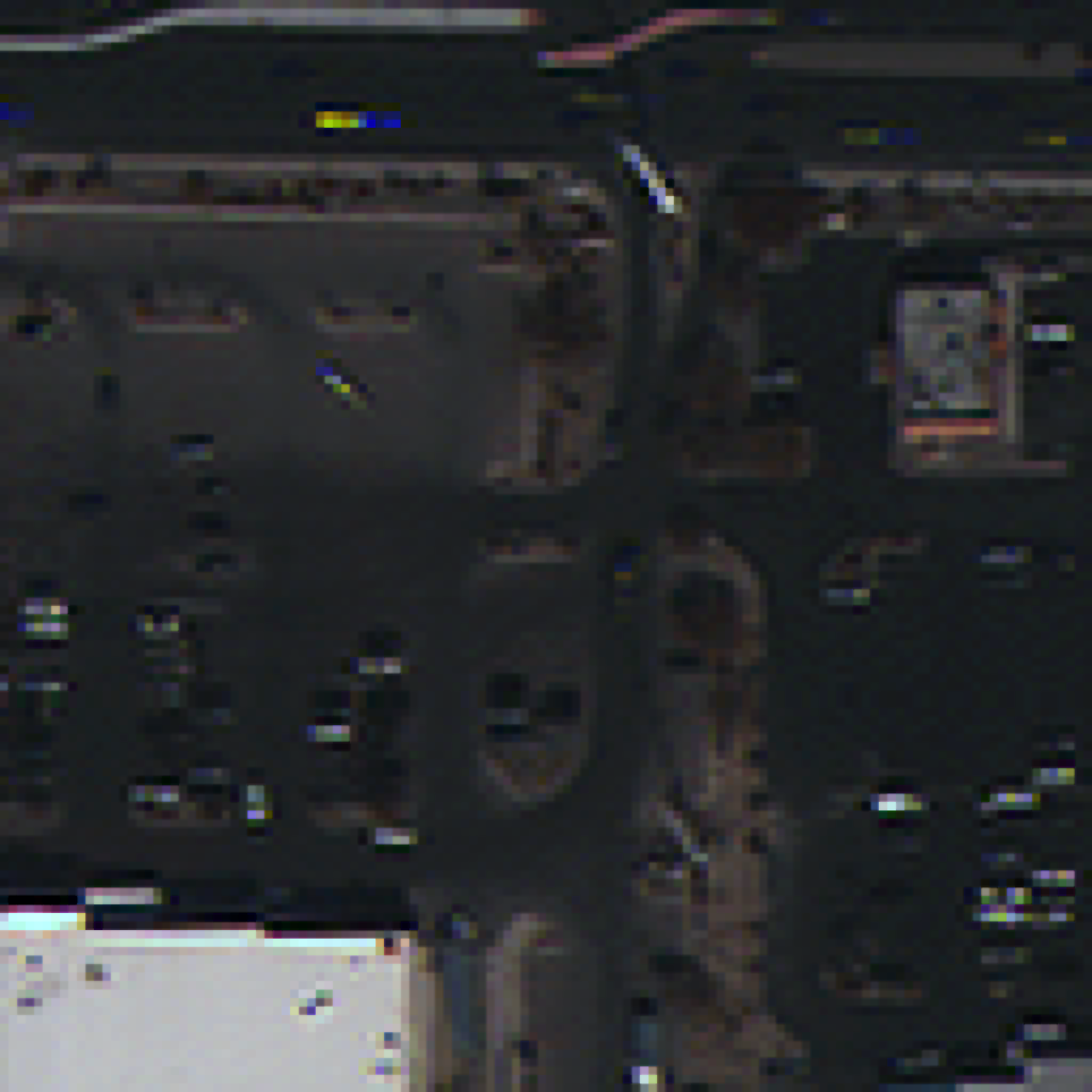} &
    \includegraphics[width=0.24\linewidth]{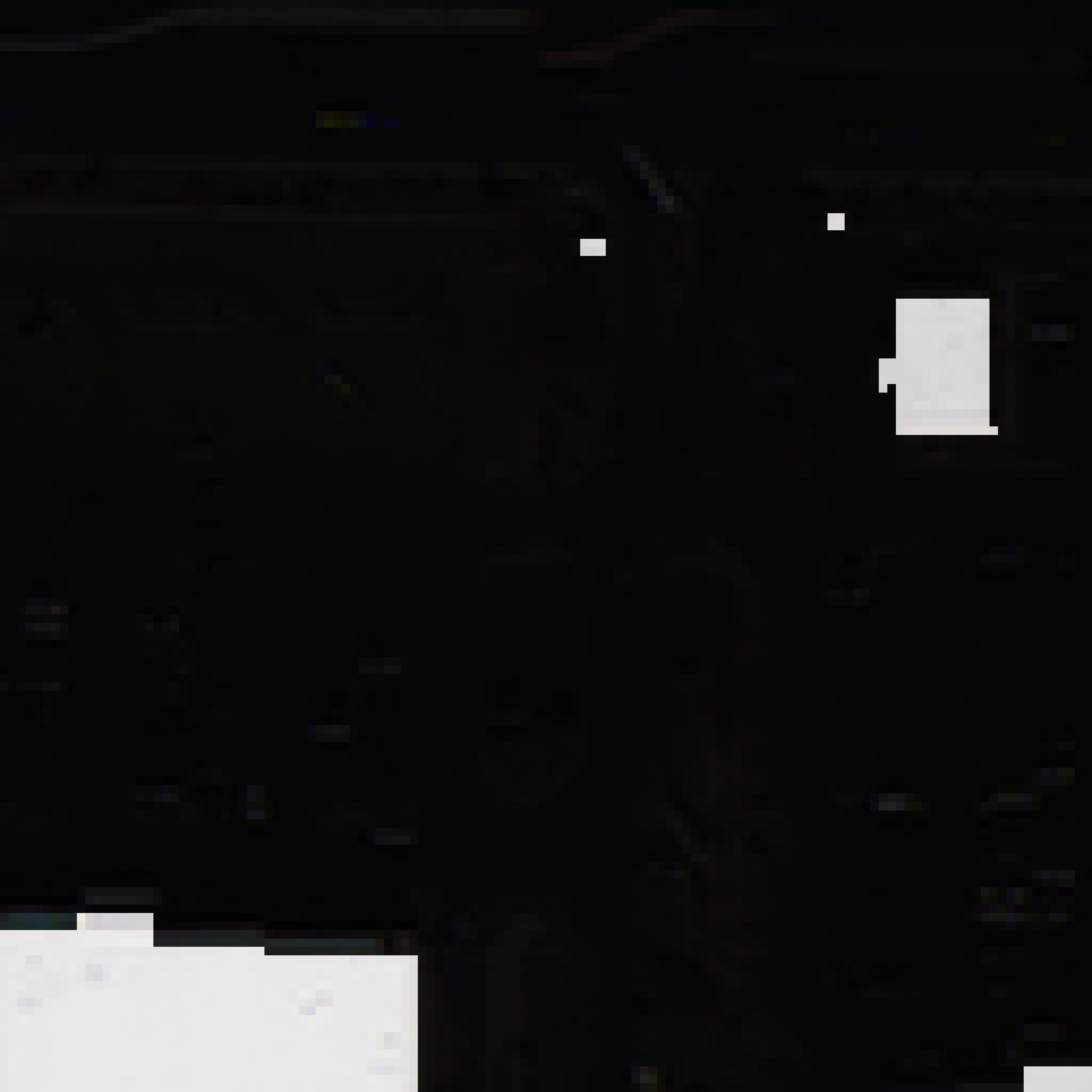} &
    \includegraphics[width=0.24\linewidth]{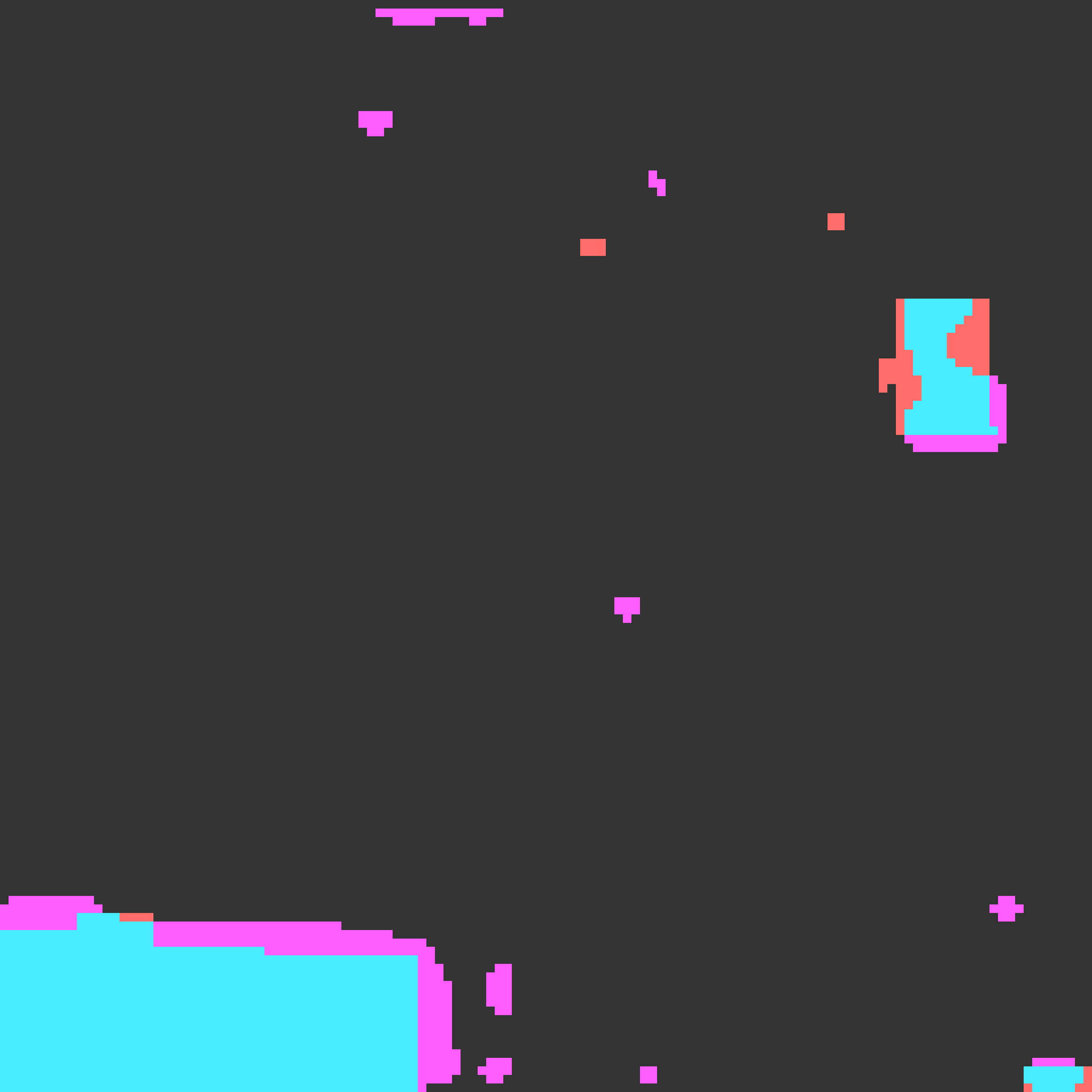} &
    \includegraphics[width=0.24\linewidth]{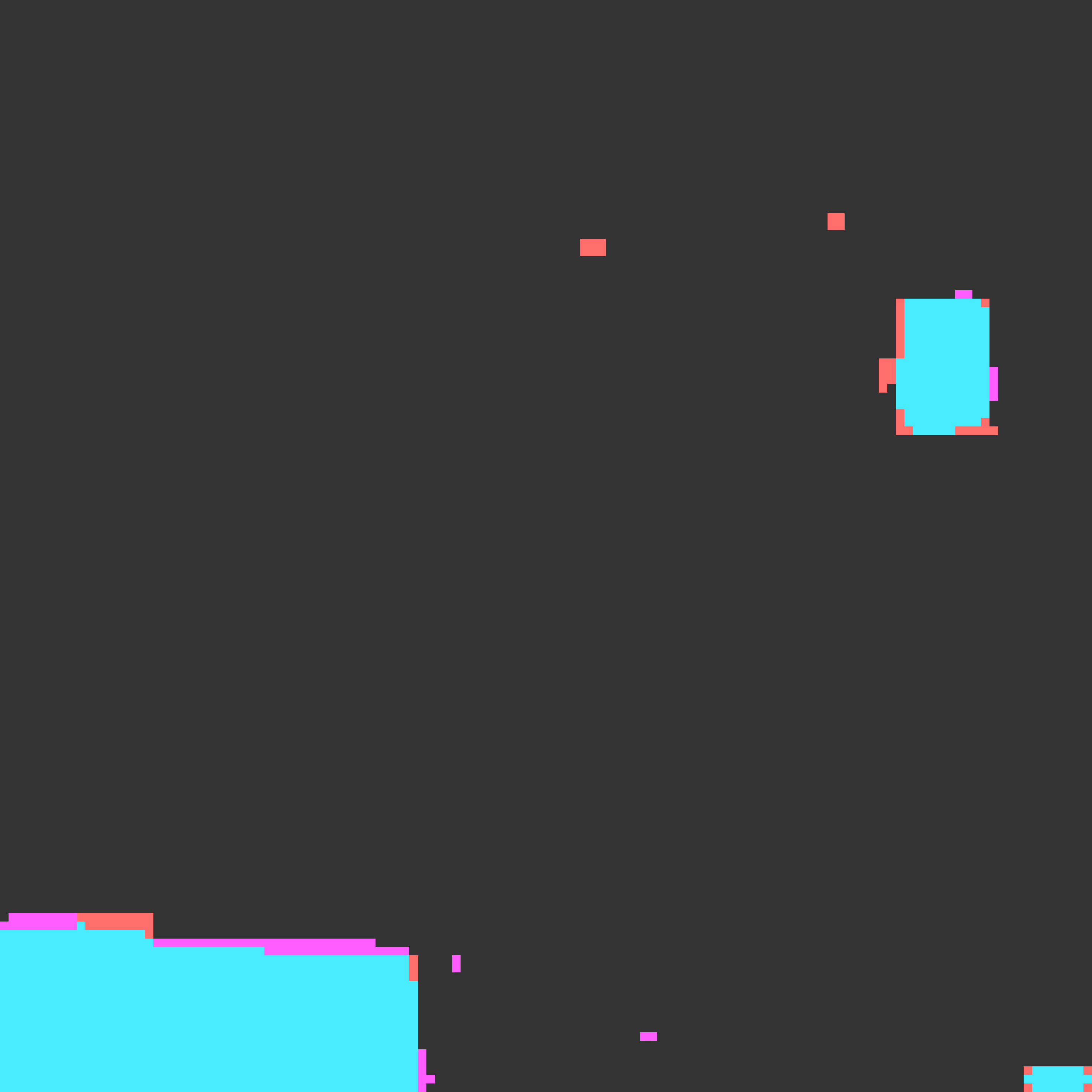} \\
    \includegraphics[width=0.24\linewidth]{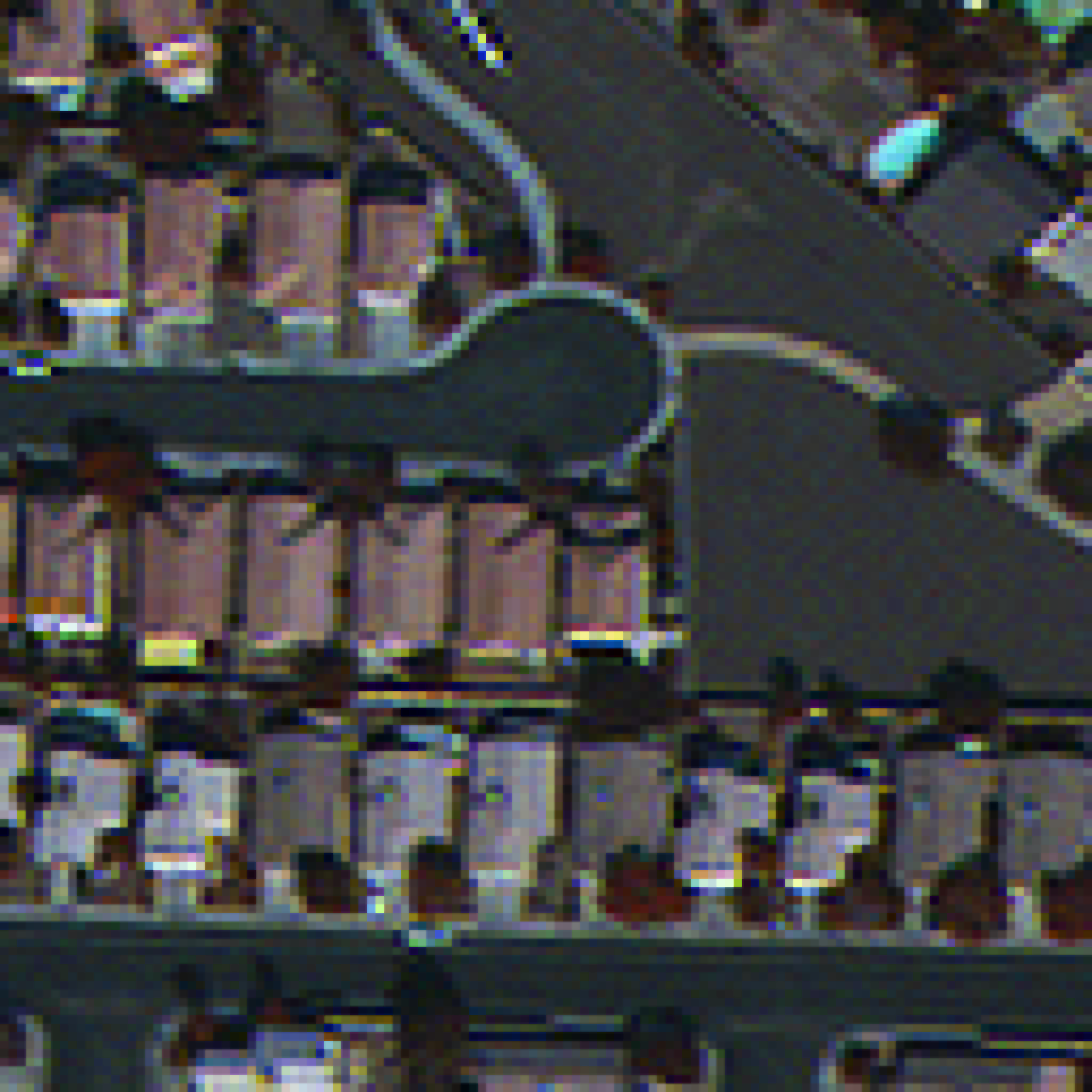} &
    \includegraphics[width=0.24\linewidth]{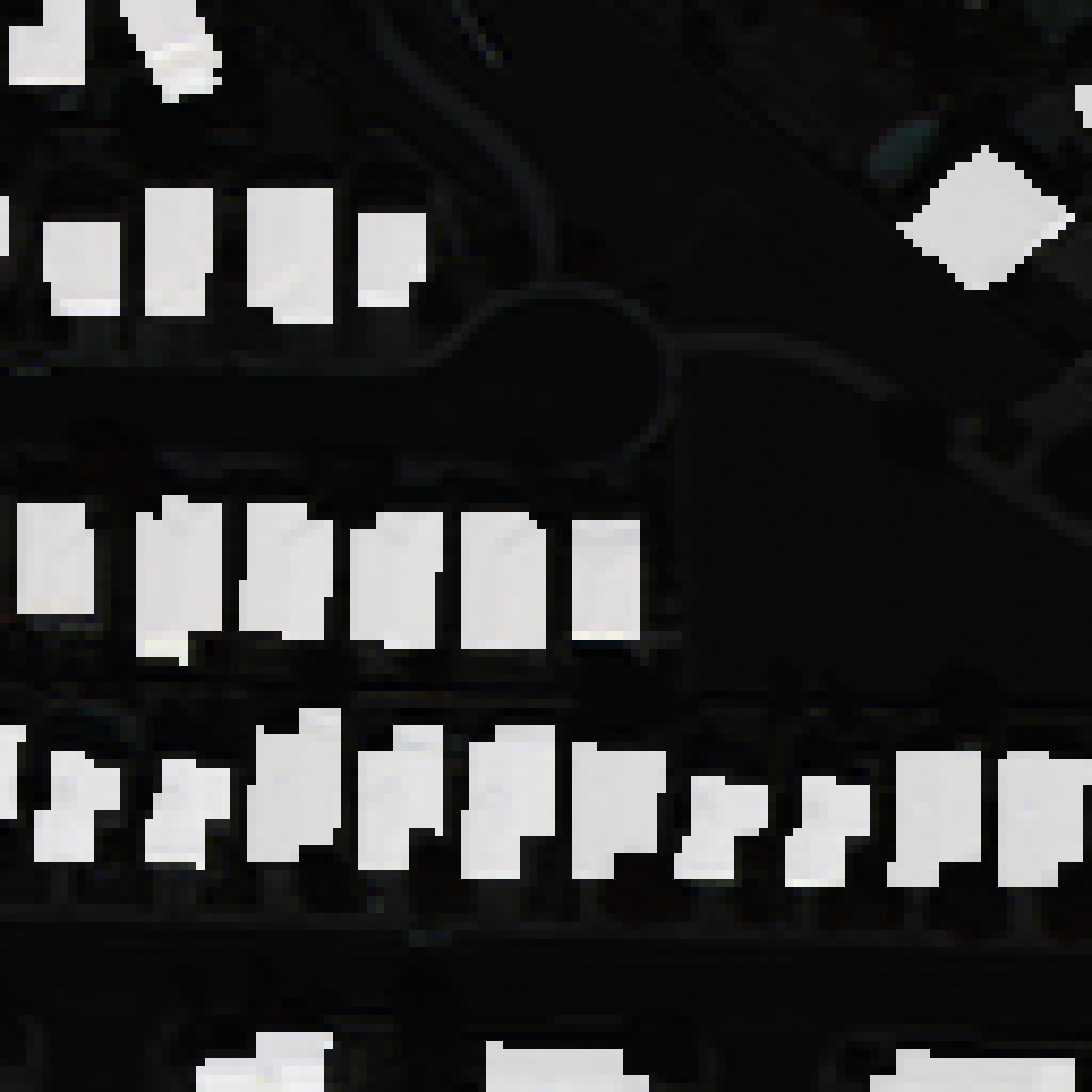} &
    \includegraphics[width=0.24\linewidth]{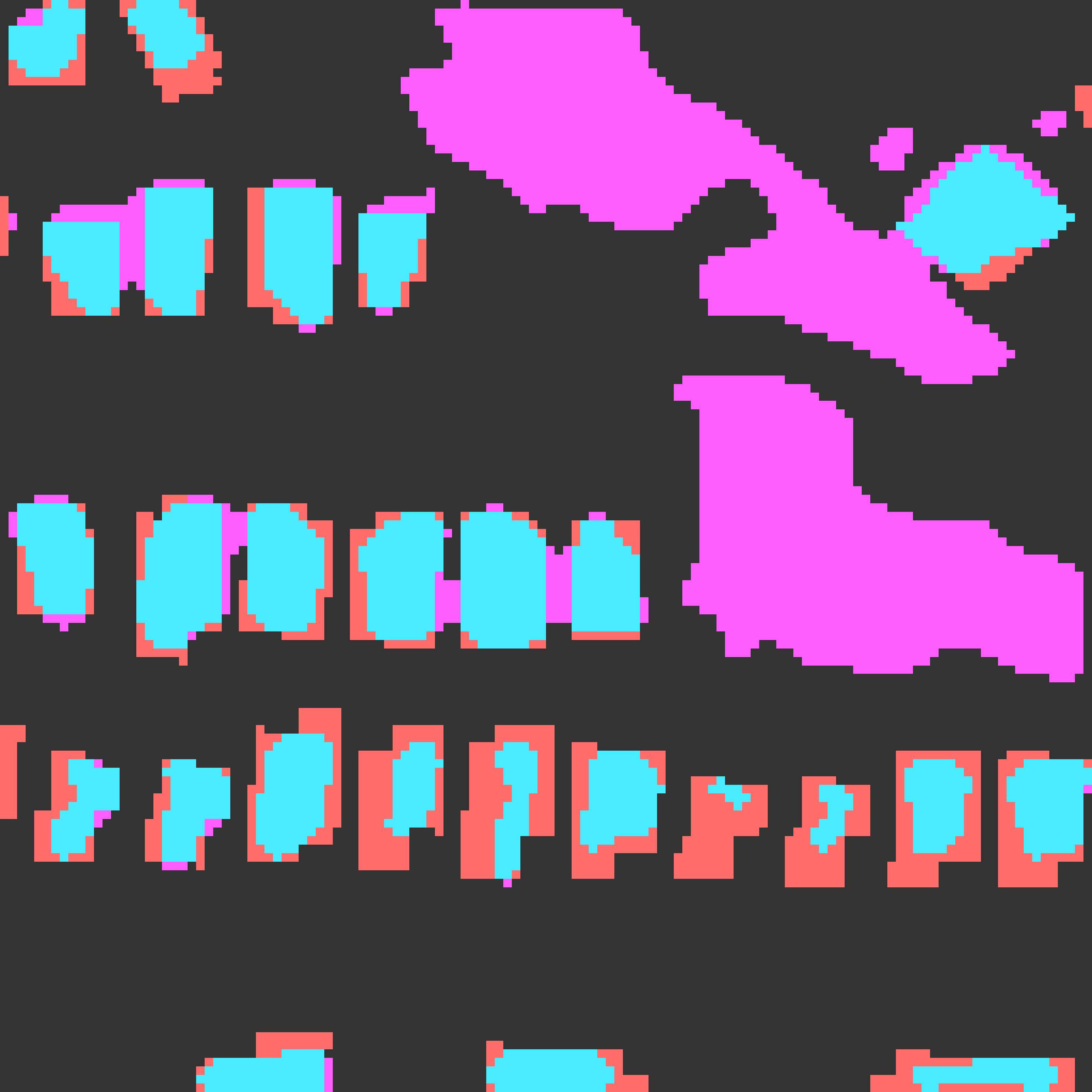} &
    \includegraphics[width=0.24\linewidth]{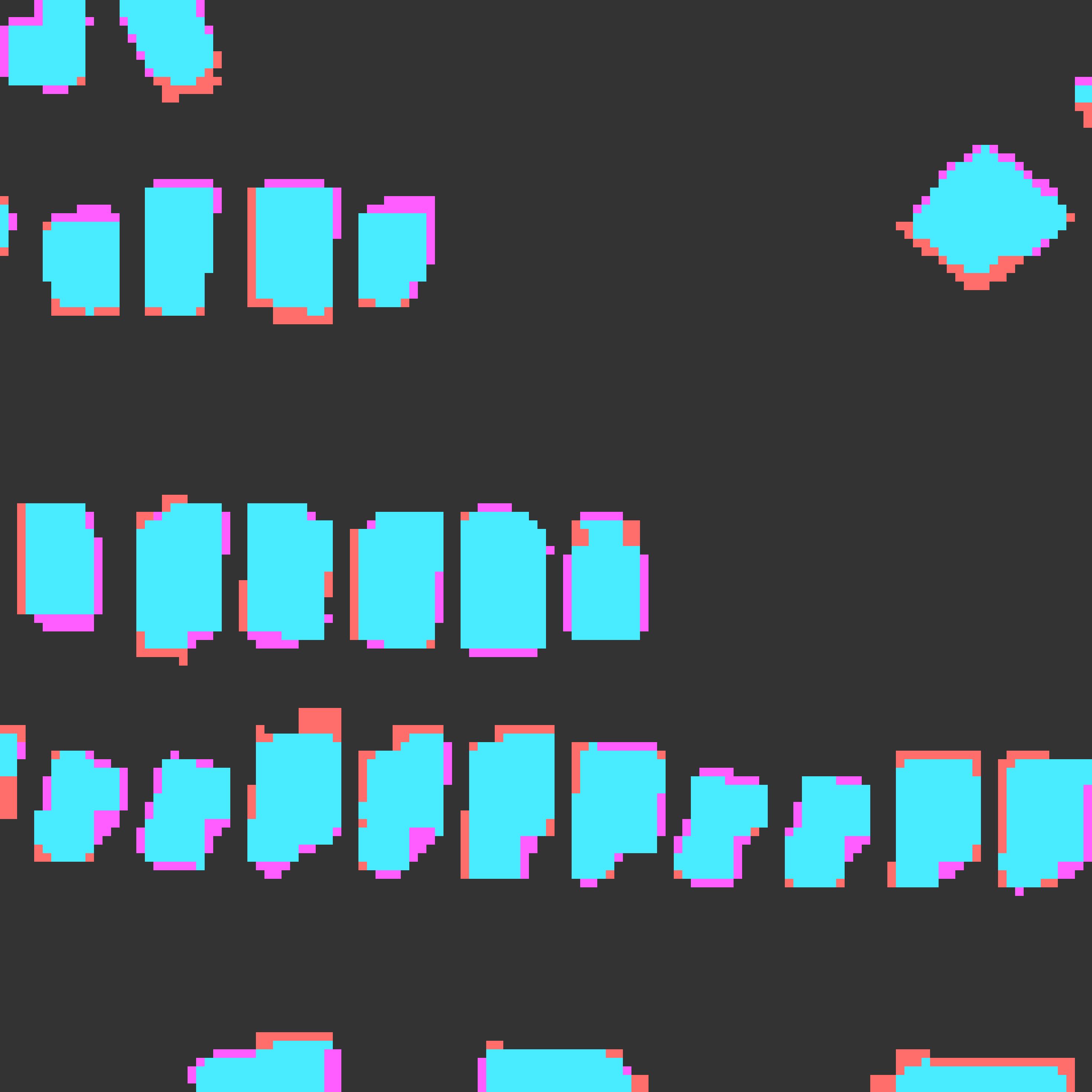} \\
    \includegraphics[width=0.24\linewidth]{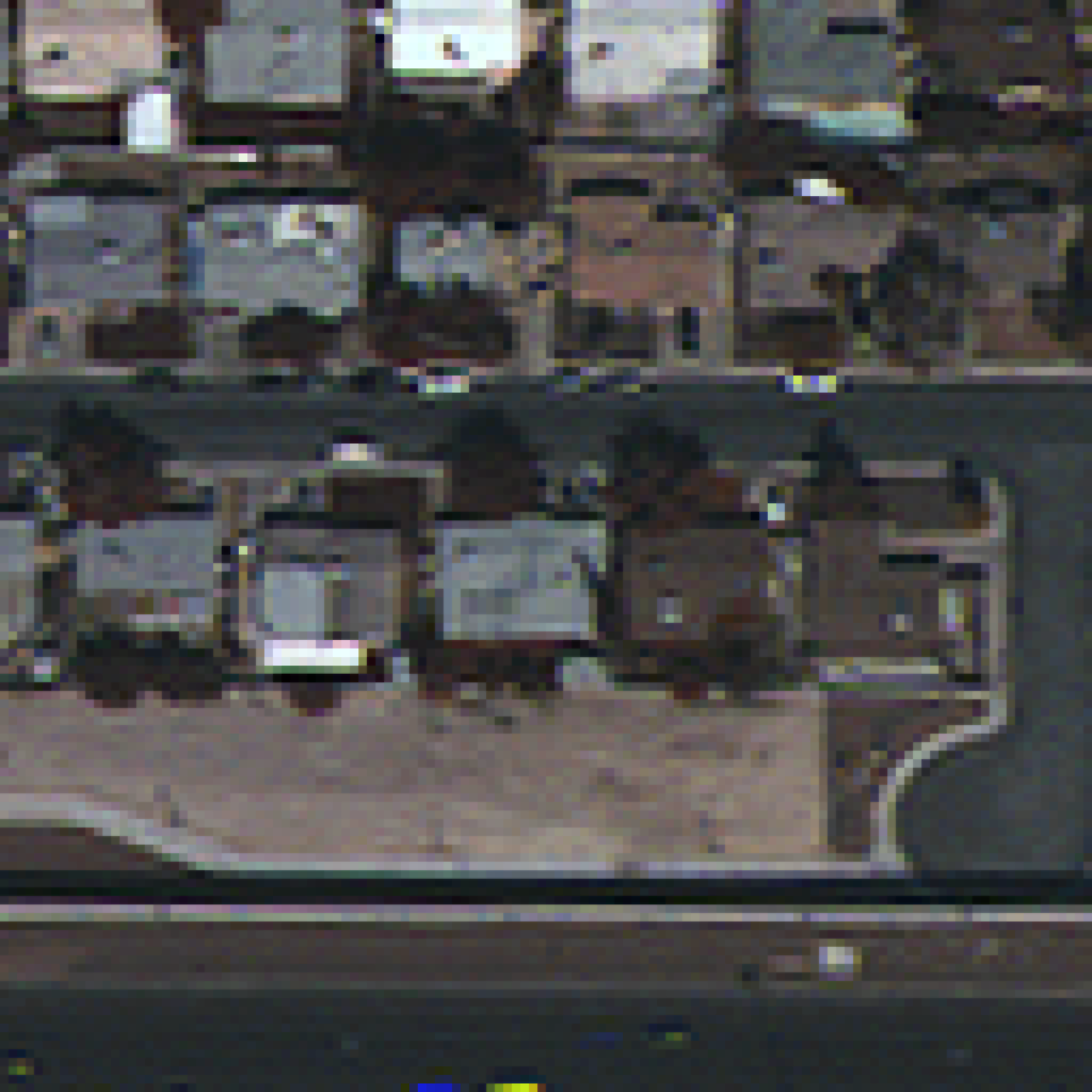} &
    \includegraphics[width=0.24\linewidth]{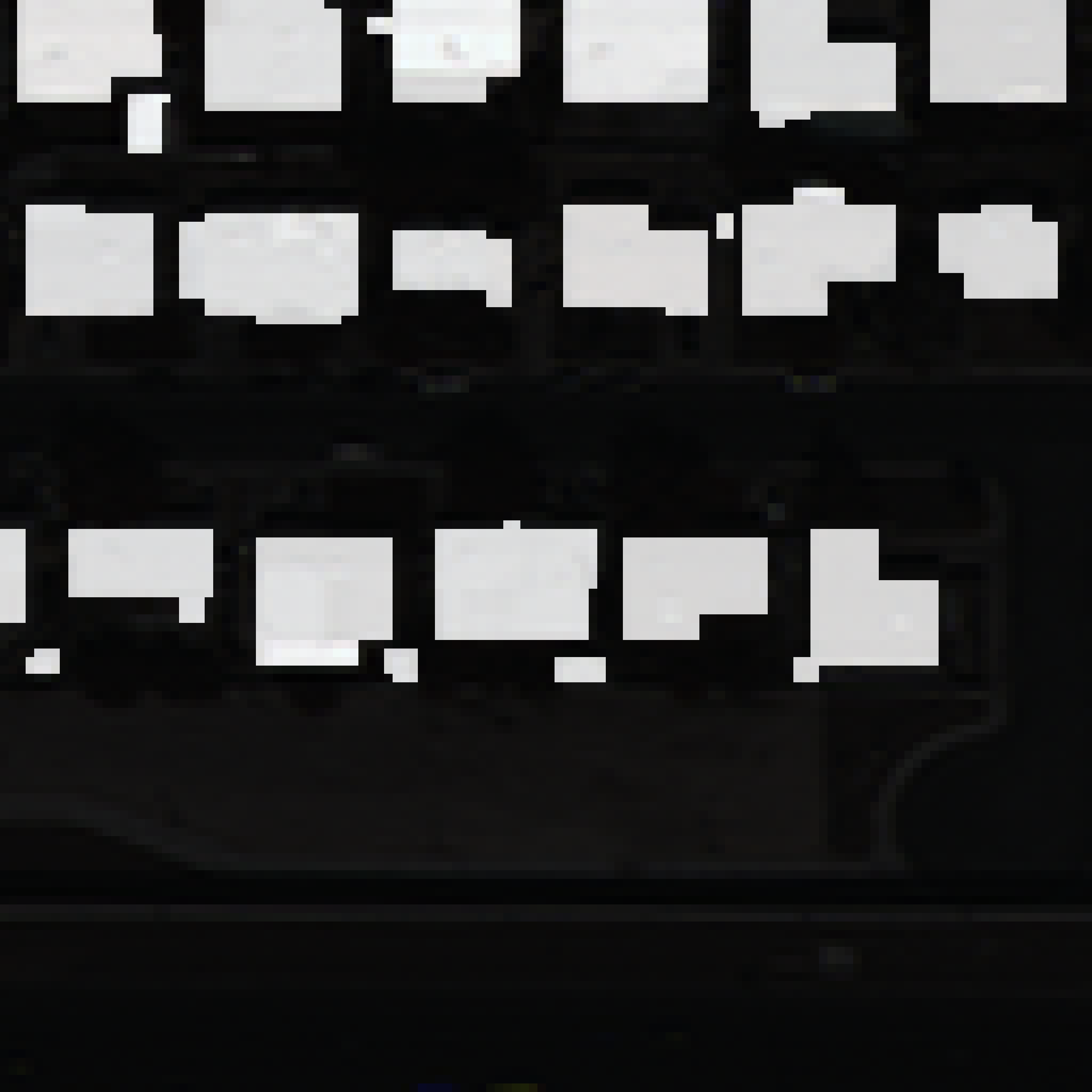} &
    \includegraphics[width=0.24\linewidth]{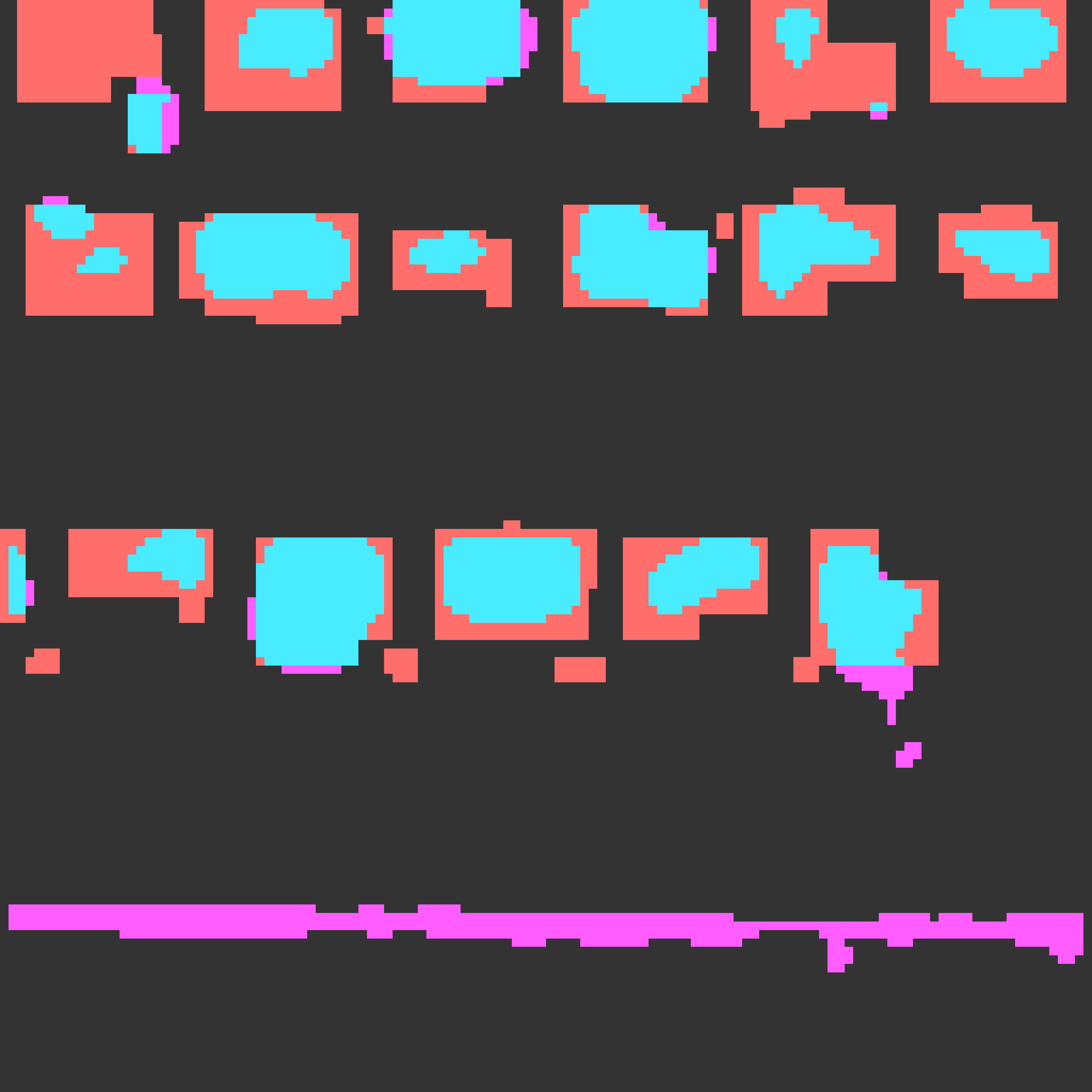} &
    \includegraphics[width=0.24\linewidth]{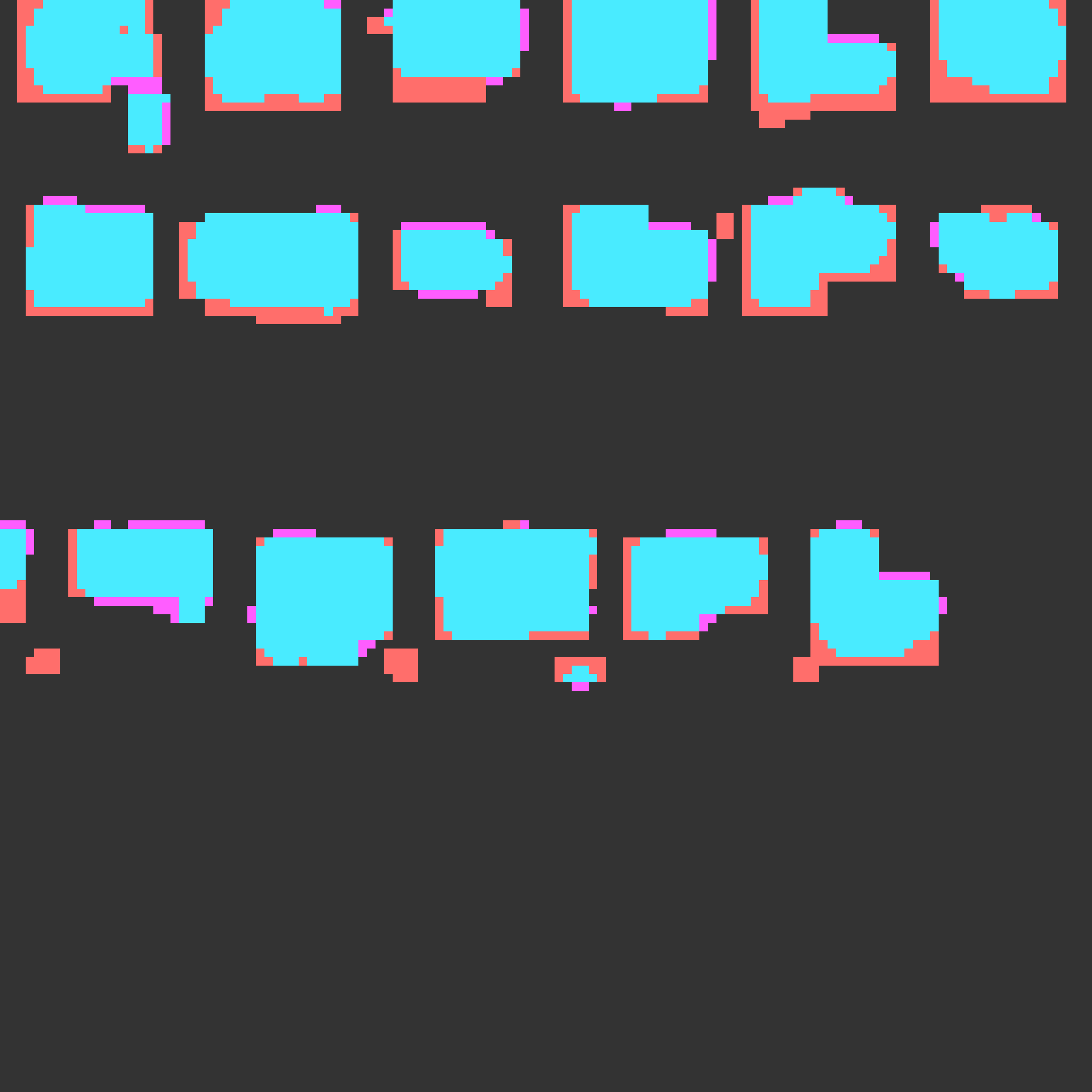} \\
    \includegraphics[width=0.24\linewidth]{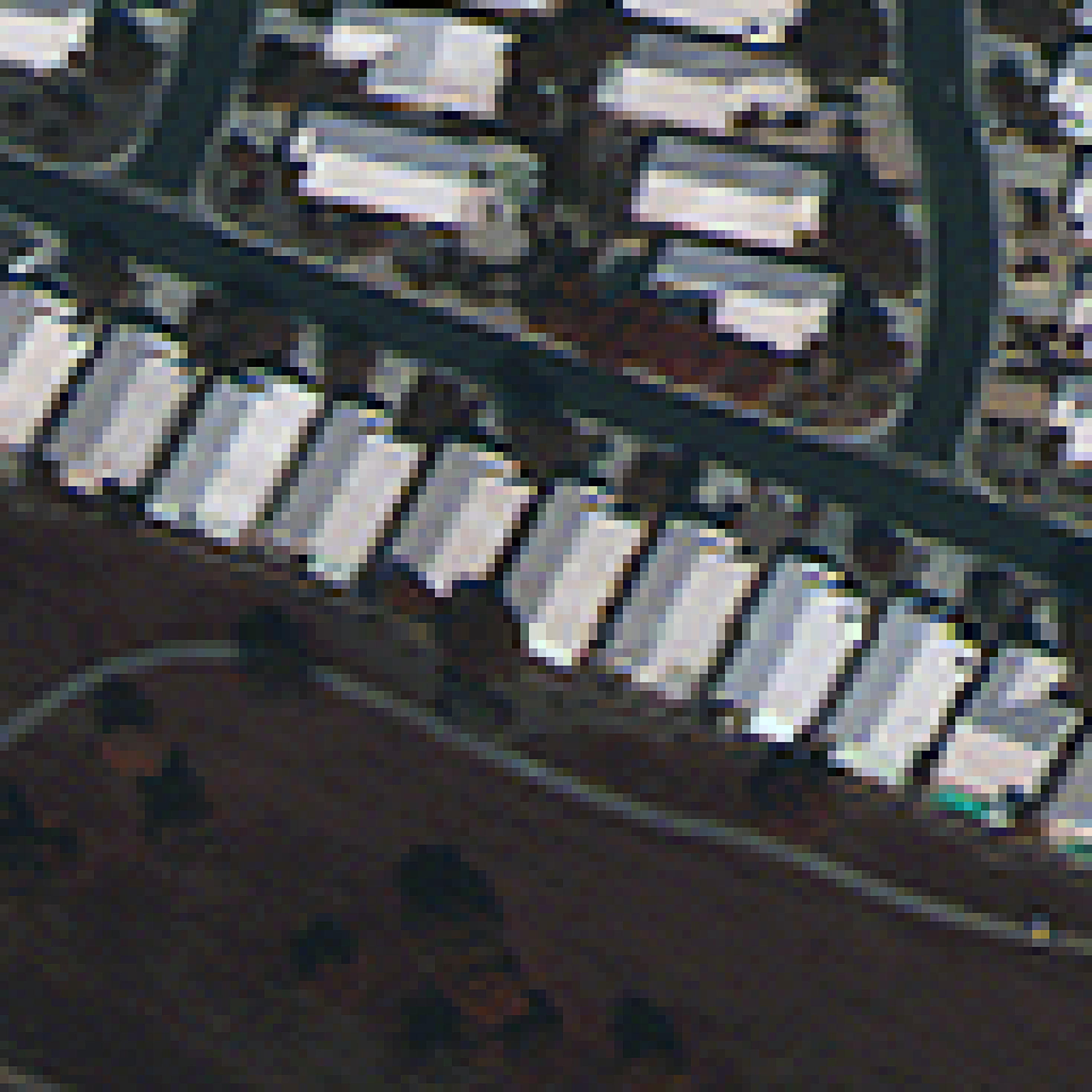} &
    \includegraphics[width=0.24\linewidth]{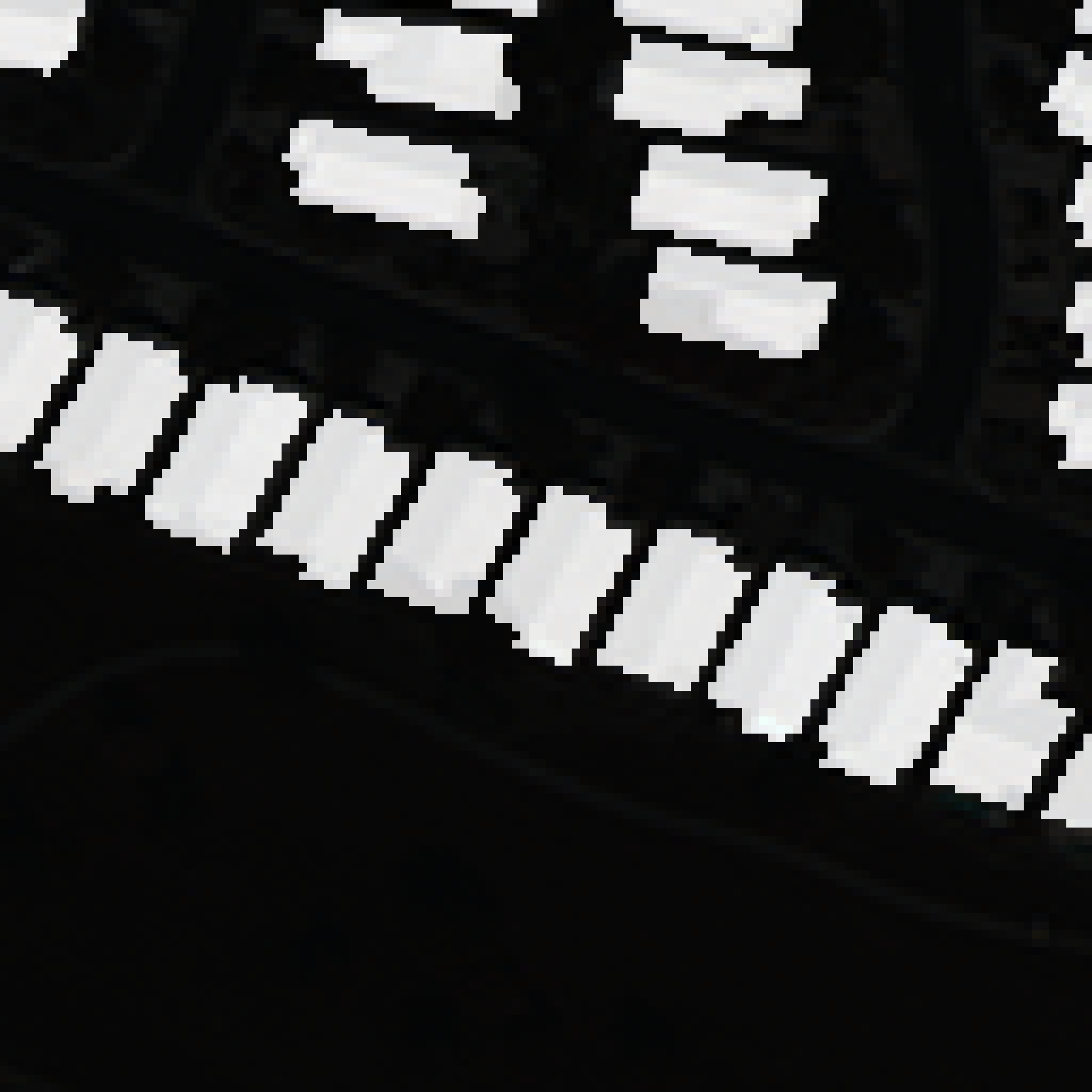} &
    \includegraphics[width=0.24\linewidth]{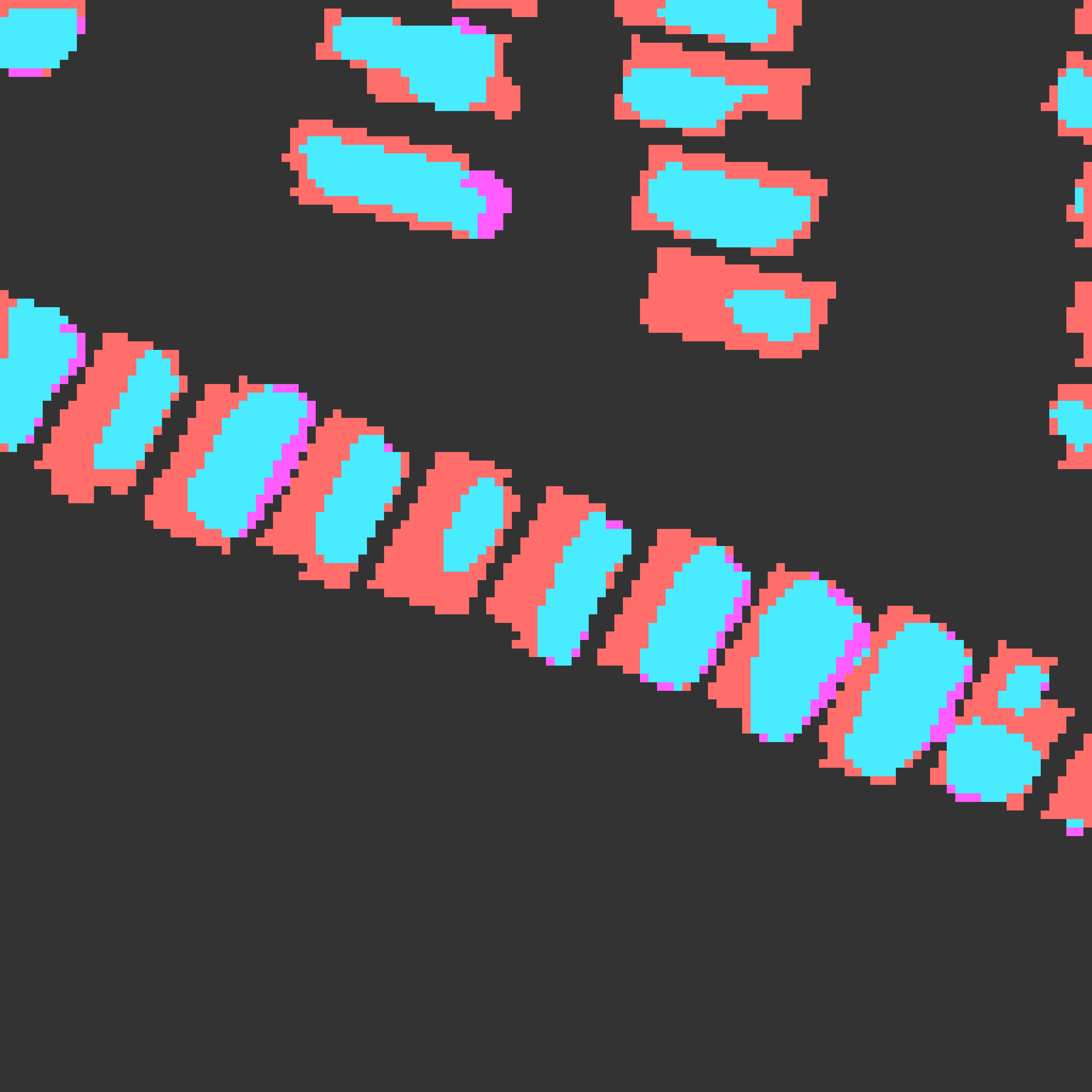} &
    \includegraphics[width=0.24\linewidth]{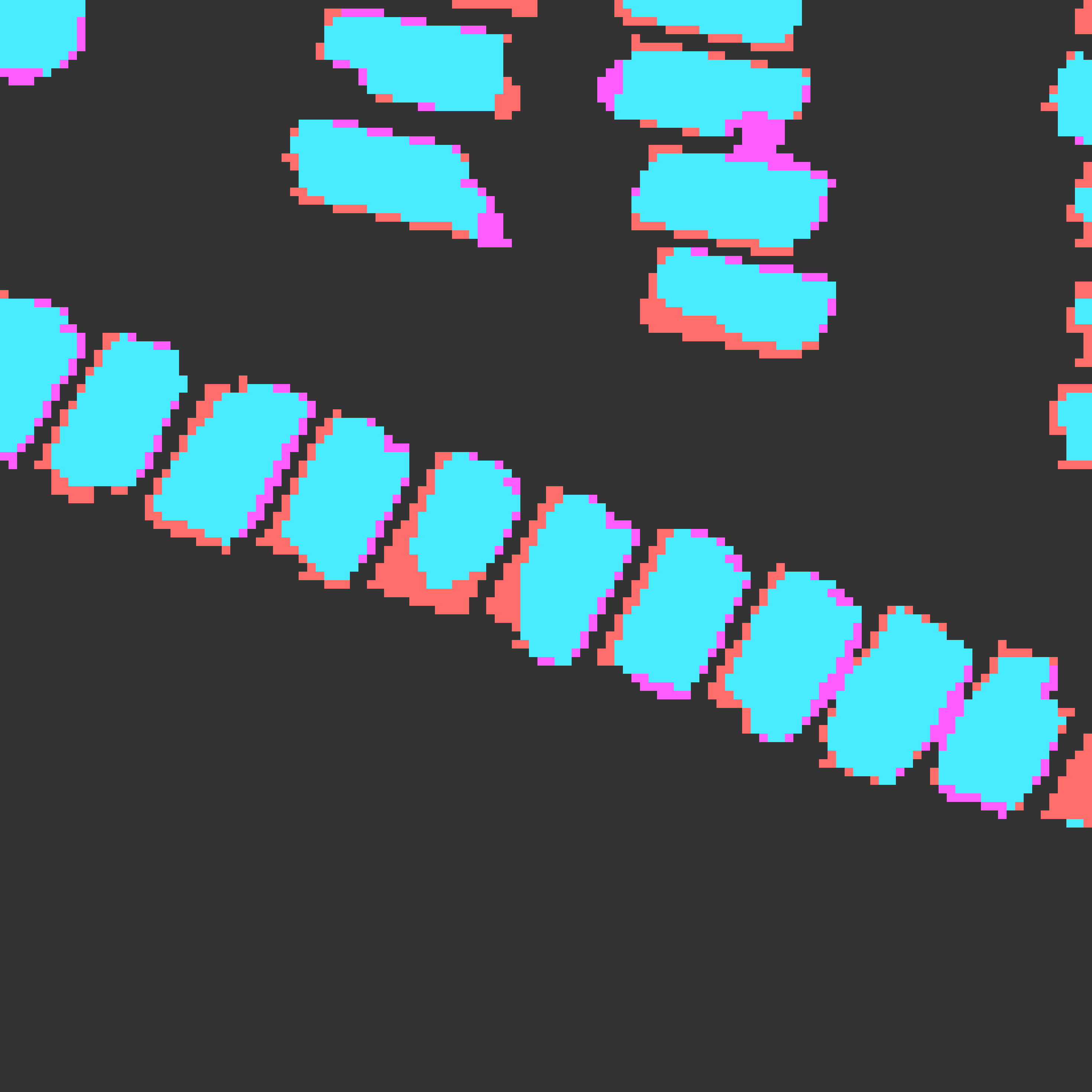} \\
    Image & 
    Ground Truth  & 
    DeepLab \cite{chen2017deeplab} &
    Ours ($\mathcal{S}+\mathcal{F}$) 
\end{tabular}
\vspace{-0.5em}
\caption{\textbf{Qualitative results} of our method on SpaceNet dataset \cite{van2018spacenet}. Cyan, magenta, gray, and red colors represent true positive, false positive, true negative, and false negative respectively. Best viewed in zoom and color.}
\vspace{-1.65em}
\label{fig:spacenet_qual_results}
\end{figure}

\subsection{Semantic Segmentation}
\vspace{-0.3em}
\paragraph{Implementation Details.} We use the SpaceNet building segmentation dataset for this task. The dataset provides 10,593 multi-spectral images of size $163 \times 163$ labeled with pixel-wise building/no-building masks, split into train and validation sets (90\%/10\%). We use a DeepLab-v3 \cite{chen2017deeplab} with skip-connections network with our frozen pre-trained backbone and fine-tune it for 24 epochs with batch-size of $32$, learning rate of $0.0085$, momentum of $0.6$, and weight decay of $0.001$.  We report mean intersection over union (mIoU) of the best performing epoch in Tab. \ref{tab:quant_results_spacenet}. The fully-supervised baselines follow the same steps as our fine-tuned approach, without the pre-trained weight initialization. 
\vspace{-1.3em}
\paragraph{Results Discussion.} Tab. \ref{tab:quant_results_spacenet} and Fig \ref{fig:spacenet_qual_results} compare the quantitative and qualitative results of baselines and our method. For our baselines, we report Ayush \etal \cite{ayush2021geography} and MoCo-v2 \cite{he2020momentum}, which use PSANet \cite{zhao2018psanet} with backbones pre-trained on geography consistency objective. We also report the performance of DeepLab-v3 initialized with random and ImageNet \cite{krizhevsky2012imagenet}  pre-trained weights. As shown in Tab. \ref{tab:quant_results_spacenet}, our method requires significantly less epochs to obtain superior performance on the SpaceNet building segmentation dataset. We outperform our self-supervised based baseline by 2.61\%, and fully supervised based baseline by 8.90\%, with 76\% convergence speed reduction.

\vspace{-0.5em}
\section{Results}
\begin{figure}[t!]
\setlength\tabcolsep{1pt}
\def\arraystretch{0.5}
\centering
\begin{tabular}{@{}cccc@{}}
    \includegraphics[width=0.23\linewidth]{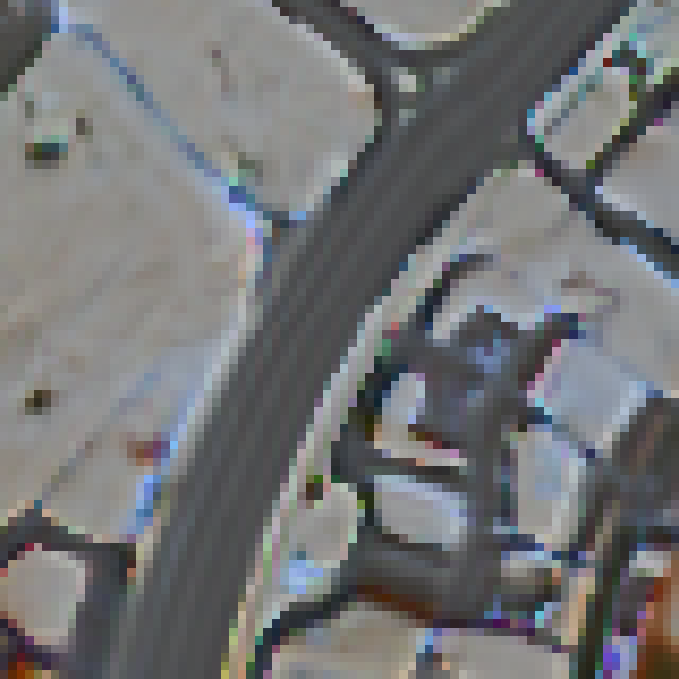} &
    \includegraphics[width=0.23\linewidth]{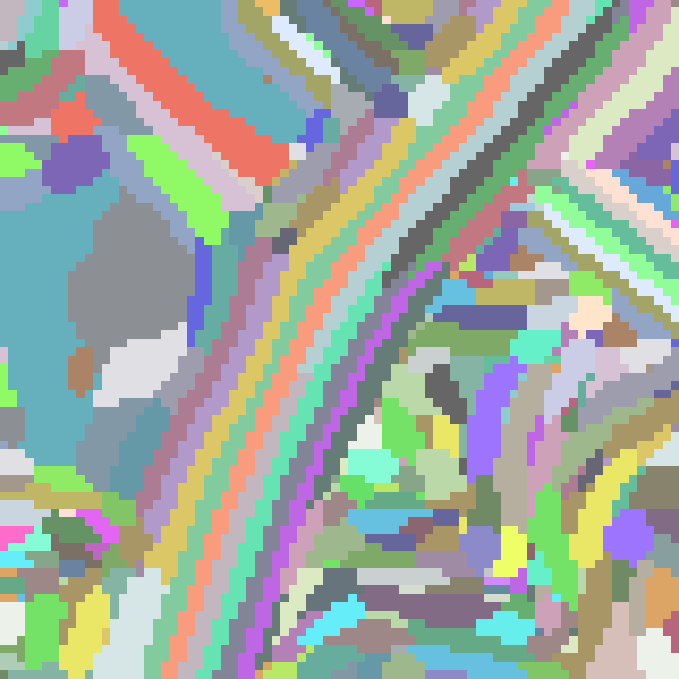} &
    \includegraphics[width=0.23\linewidth]{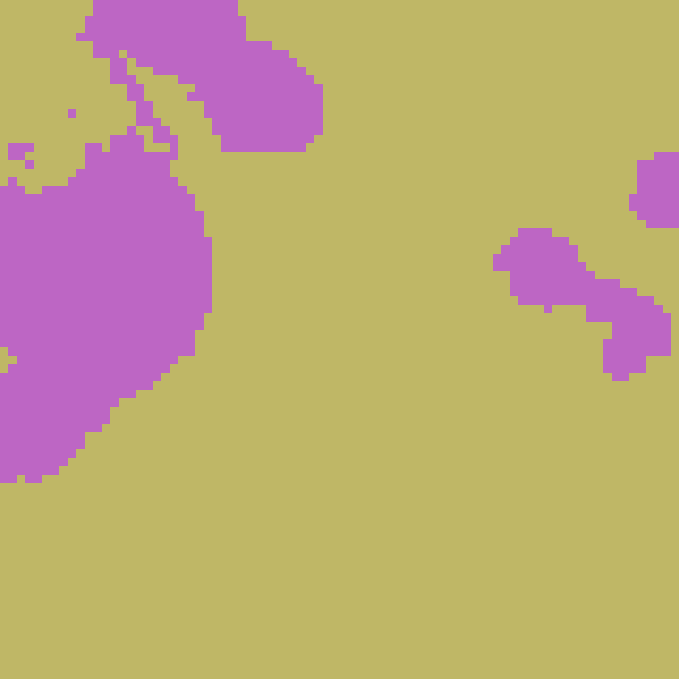} &
    \includegraphics[width=0.23\linewidth]{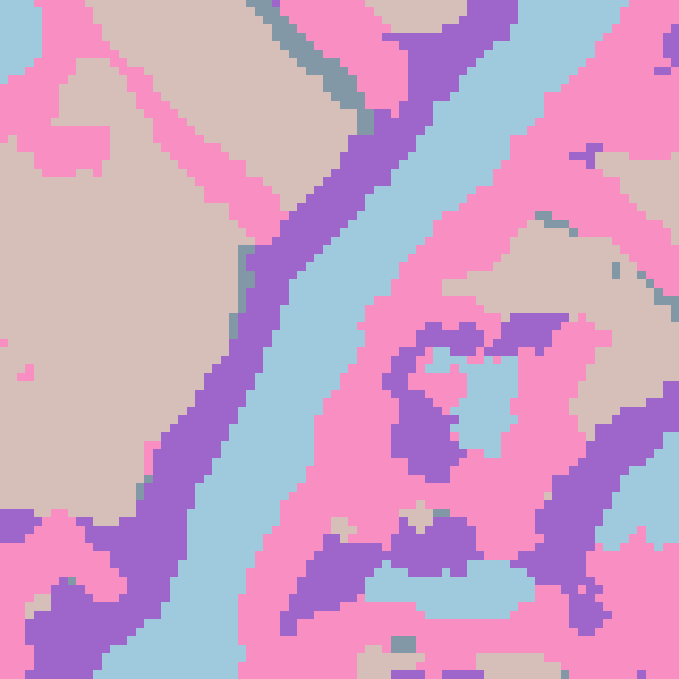} \\
    \includegraphics[width=0.23\linewidth]{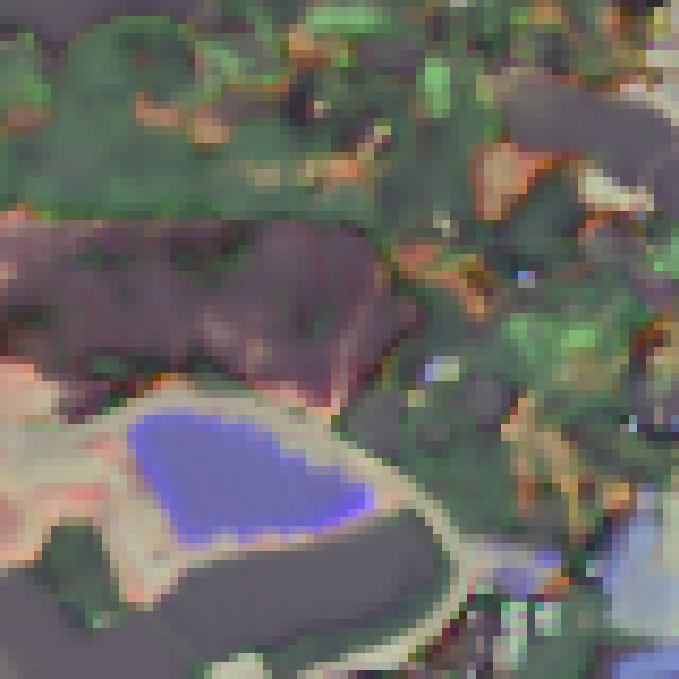} &
    \includegraphics[width=0.23\linewidth]{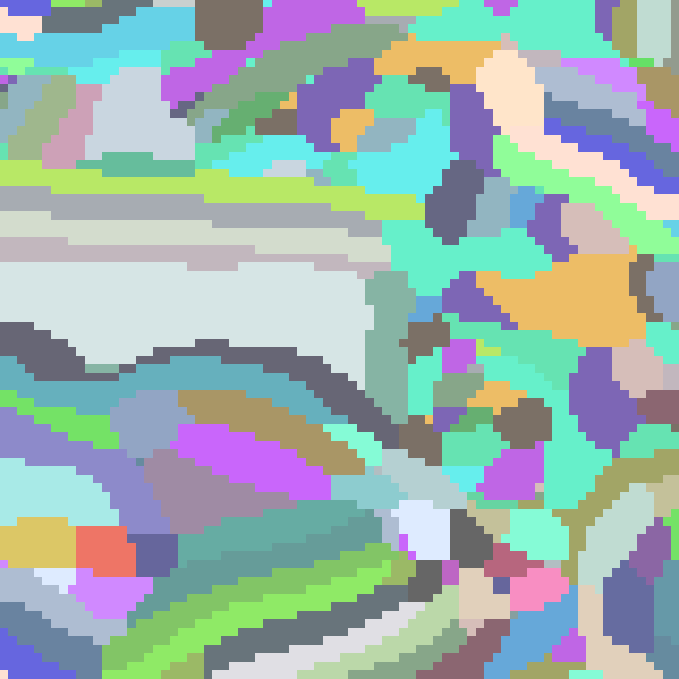} &
    \includegraphics[width=0.23\linewidth]{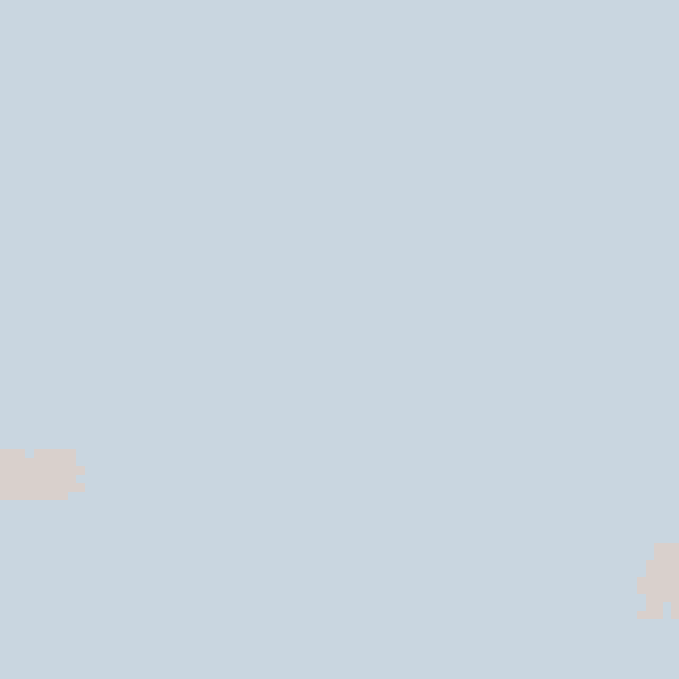} &
    \includegraphics[width=0.23\linewidth]{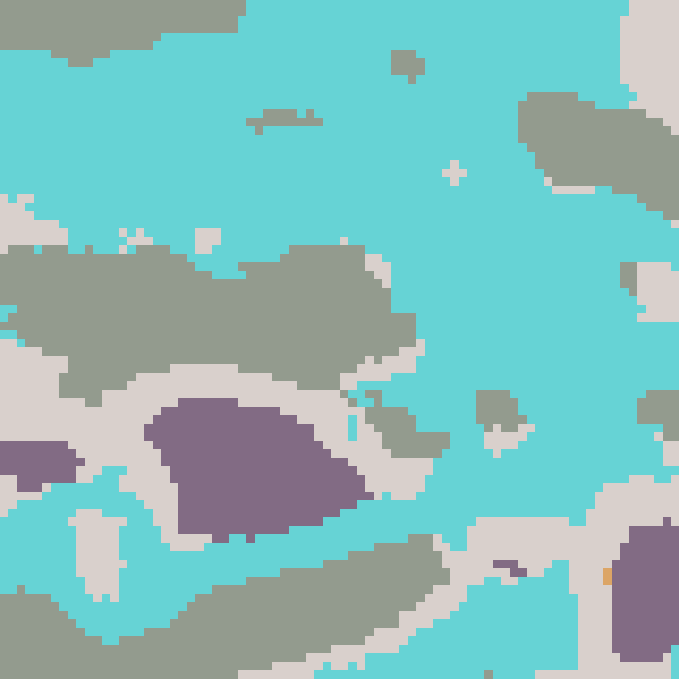} \\
    Image & 
    Textons &
    \multicolumn{1}{m{2.0cm}}{\centering Patch-wise Backbone} &
    Ours 
\end{tabular}
\vspace{-0.7em}
\caption{\textbf{Qualitative evaluation} of our generated material and texture based visual word maps. It can be seen that our method provides more descriptive surface-based features that are not highly sensitive to small texture irregularities like textons, or under-sensitive to structure changes like the patch-wise backbone. Best viewed in zoom and color. Colors are random.}
\vspace{-0.9em}
\label{fig:visual_words}
\end{figure}

\vspace{-0.3em}
Evident by our qualitative and quantitative results, our method provides both superior performance and convergence time (measured in epochs) for the evaluated downstream tasks. It is shown that material and texture are strongly associated with common remote sensing downstream tasks, and the ability to represent material and texture effectively improves performances on those tasks. Since quantitatively measuring the ability to represent material without material labels is difficult, we analyze and showcase qualitative texture and material results in the form of visual word maps (pixel-wise cluster assignments). We also discuss limitations, running time, pseudo-code, and additional qualitative results in the supplementary material.
%
%
%
\vspace{-1.2em}
\paragraph{Visual Word Maps Generation.} In order to measure the effectiveness of our approach to describe materials and textures, we qualitatively evaluate the visual word maps (pixel-wise cluster assignments) generated by our method. Ideally, we expect similar material and textures to be mapped to the same clusters, without over or under grouping of pixels. We visually compare classical textons, a patch-wise backbone, and our method in Fig. \ref{fig:visual_words}. The patch-wise backbone has the same base architecture as MATTER, but without TeRN and surface residual encoding modules. Both methods were trained on the same dataset, with the same hyperparameters, and number of iterations, as described in Sec. \ref{sec:pre_training_subsec}. It can be seen that the textons and patch-wise backbone approaches generate two extreme cases of over-sensitivity and under-sensitivity to changes in material and texture. Since textons operate on raw intensity values, the inter-material variance is small, making it highly sensitive to small texture changes. This can be seen in the textons-generated visual word map, in which small irregularities on the road results in mapping to different visual words. On the other hand, the patch-wise backbone, even with the receptive field constraints through patched inputs, still loses crucial low-level details essential for texture representation. This is indicated by the grouping of obviously different textures to a single visual word. In contrast, as demonstrated in Fig. \ref{fig:visual_words}, our Texture Refinement Network and Surface Residual Encoder boost the impact of low-level features, generating surface-based visual word maps. Our method is able to retain texture-essential features, and generalize surface representation which translates to superior surface-based visual word maps. 
\begin{table}[t!]
\centering
\resizebox{0.47\textwidth}{!}{%
\begin{tabular}{c|ccccccccc}
\toprule
\multicolumn{10}{c}{Inference Crop Size} \\ 
 \cmidrule(lr{1em}){2-10}
 \multirow{9}{*}{\rotatebox[origin=c]{90}{Train Crop Size}}
& -  & 7 & 9 & 11 & 13 & 15 & 17 & 19 & 21 \\ 
& 5  & 46.68 & 47.38 & 46.94 & 46.94 & 45.74 & 44.51 & 43.74 & 42.95 \\
& 7  & 48.52 & \textbf{49.48} & 49.01 & 49.02 & 47.76 & 46.64 & 45.92 & 44.69 \\
& 9  & 48.58 & 47.60 & 48.02 & 47.83 & 46.51 & 45.57 & 45.45 & 43.27 \\
& 11  & 48.98 & 47.83 & 47.32 & 46.65 & 45.64 & 44.51 & 44.16 & 42.45 \\
& 13  & 47.46 & 47.14 & 46.35 & 46.99 & 44.65 & 43.79 & 43.09 & 41.61 \\
& 15  & 47.63 & 47.15 & 47.30 & 46.10 & 45.55 & 44.68 & 44.10 & 41.85 \\
& 17  & 46.74 & 46.81 & 46.49 & 45.92 & 44.69 & 43.60 & 43.19 & 41.09 \\
\bottomrule
\end{tabular}%
}
\vspace{-0.8em}
\caption{\textbf{Receptive Field Constraint Analysis}. F-1 score (\%) performance for the unsupervised change detection task. Reported values are of the ``change'' class with respect to training and inference crop sizes (without fine-tuning). It can be seen that the method benefits from smaller receptive field, achieving superior performance when using smaller train and inference crop sizes.}
\vspace{-1.7em}
\label{tab:ablation_receptive_field}
\end{table}

\vspace{-0.35em}
\subsection{Ablation Study}
\label{sec:ablation}
\vspace{-0.35em}
\paragraph{Constraining Receptive Field.} In Tab. \ref{tab:ablation_receptive_field} we study the effects of varying receptive field constraints on our method. As mentioned before, as the receptive field increases, the impact of low-level features diminishes, along with the quality of material and texture representation. Unlike traditional methods, which resort to the usage of smaller networks to reduce receptive field, we explicitly constrain the method by feeding crops to the network, removing them from any global context. Recall that the objective of our method is to learn representation of material and spatial distribution of micro-structures, which are largely affected by low level features which are diminished in larger receptive field methods. In practice, the largest possible receptive field of our network during training is $7 \times 7 = 49$ pixels, which is significantly smaller than the receptive fields of ResNet-50, and ResNet-101, and ResNet-152 with sizes of 483, 1027, and 1507 pixels respectively. It can be seen in Tab. \ref{tab:ablation_receptive_field} that in fact, the method benefits from removal from global spatial context and smaller receptive fields, helping it learn better representation for material and texture and achieve better performance on the unsupervised change detection task. Our best results are achieved with train crop size of $7 \times 7$, and inference crop size of $9 \times 9$, while worst performance is achieved with largest training and inference receptive fields.
%
\vspace{-1.3em}
\paragraph{Impact of Modules.} In Tab. \ref{tab:ablation} we study the impact of each module in our proposed method. We evaluate performance of the self-supervised change detection task (F-1 score of ``change" class) as an ablation metric since it has strong transferability to material and texture representation learning. We consider all possible network combinations with patch-wise backbone, TeRN, and Surface Residual encoder. The patch-wise backbone corresponds to the network fed by patch-wise inputs, without TeRN or Surface Residual encoder. We then selectively add TeRN and Surface Residual Encoder to the network and record its performance. Every network combination was trained and evaluated with the same hyperparameters and procedure described in Sec. \ref{sec:pre_training_subsec} and \ref{sec:change_detection}. It can be seen that each module provides incremental performance boost, with best performance achieved when both modules are implemented in the network.


\begin{table}[t!]
\centering
\resizebox{0.47\textwidth}{!}{%
\begin{tabular}{@{}c@{\hspace{0.68em}}c@{\hspace{0.68em}}c@{\hspace{0.68em}}c@{}}
\toprule
Patch-wise Backbone & Texture Refinement & Surface Residual & F-1 Score (\%)\\ \midrule
$\checkmark$ &   &  &  37.42\\ 
$\checkmark$ &   & $\checkmark$ & 41.84 \\ 
$\checkmark$ &  $\checkmark$ &   & 43.23 \\ 
$\checkmark$ & $\checkmark$ & $\checkmark$ &  \textbf{49.48} \\ 
\bottomrule
\end{tabular}%
}
\vspace{-0.8 em}
\caption{\textbf{Ablation study.} F-1 score of the ``change" class of the Onera Satellite Change Detection dataset using the self-supervised approach with respect to modules used.  }
\vspace{-1.8em}
\label{tab:ablation}
\end{table}

\vspace{-0.8em}
\section{Conclusion}
\vspace{-0.4em}

In this work, we present MATTER, a novel self-supervised method that learns material and texture based representation for multi-temporal, spatially aligned satellite imagery. By utilizing patch-wise inputs and our refinement network, we constrain the receptive field and enhance texture-essential features. Those are then mapped to residuals of learned clusters as an affinity measurement, which represents the material and texture composition of the sampled patch. Through our self-supervision pipeline, MATTER learns discriminative features for various material and texture surfaces, which are shown to have strong correlation to change (change of surface infers actual change), or can be used as pre-trained weights for other remote sensing tasks.


{\small
\bibliographystyle{ieee_fullname}
\bibliography{egbib}
}

\end{document}